\documentclass[11pt]{article}

\usepackage[final]{acl}

\usepackage{times}
\usepackage{latexsym}

\usepackage[T1]{fontenc}

\usepackage[utf8]{inputenc}

\usepackage{microtype}

\usepackage{inconsolata}

\usepackage{graphicx}

\usepackage{hyperref}       
\usepackage{url}            
\usepackage{booktabs}       
\usepackage{amsfonts}       
\usepackage{nicefrac}       
\usepackage{xcolor}         
\usepackage{algorithm}
\usepackage{algpseudocode}
\usepackage{subfigure}
\usepackage{tikz}
\usetikzlibrary{shapes,arrows,arrows.meta,positioning,fit,backgrounds}
\usepackage{amsmath}
\usepackage{pgfplots}  
\usepackage{pgfplotstable}
\pgfplotsset{compat=1.18}
\usepackage{tcolorbox}
\tcbuselibrary{skins, breakable}
\usepackage{verbatim}  
\usepackage{caption}
\usepackage{enumitem}
\usepackage{fontawesome5}
\usepackage{xspace}
\usepackage{multirow} 
\usepackage{algpseudocodex} 
\usepackage{titletoc}
\usepackage{listings}
\usepackage[table]{xcolor}
\usepgfplotslibrary{groupplots} 
\definecolor{softgray}{gray}{0.95}
\usetikzlibrary{plotmarks}
\usepackage{tabularx}
\usepackage{array}
\usepackage{makecell}
\usepackage{seqsplit}
\usepackage{dblfloatfix}
\usepackage{placeins}

\newcommand{\socia}{\texttt{SOCIA-EVO}\xspace}
\newcommand{\ofsocia}{\texttt{SOCIA-EVO}’s\xspace}

\title{\socia: Automated Simulator Construction via \\ Dual-Anchored Bi-Level Optimization}

\author{\textbf{Yuncheng Hua},~\textbf{Sion Weatherhead},~\textbf{Mehdi Jafari},~\textbf{Hao Xue},~\textbf{Flora D. Salim}\footnotemark[2]\\
School of Computer Science and Engineering, University of New South Wales, Australia\\
\{devin.hua, s.weatherhead, mahdi.jafari, hao.xue1, flora.salim\}@unsw.edu.au\\ 
}

\begin{document}
\maketitle

\renewcommand{\thefootnote}{\fnsymbol{footnote}}
\footnotetext[2]{Corresponding author.}

\begin{abstract}
Automated simulator construction requires \emph{distributional fidelity}, distinguishing it from generic code generation. We identify two failure modes in long-horizon LLM agents: \emph{contextual drift} and \emph{optimization instability} arising from conflating structural and parametric errors. We propose \textbf{\socia}, a dual-anchored evolutionary framework. \socia introduces (1) a static \emph{blueprint} to enforce empirical constraints; (2) a \emph{bi-level optimization} to decouple structural refinement from parameter calibration; and (3) a \emph{self-curating Strategy Playbook} that manages remedial hypotheses via Bayesian-weighted retrieval. By falsifying ineffective strategies through execution feedback, \socia achieves robust convergence, generating simulators that are statistically consistent with observational data. 
\ofsocia code and data are available here: \url{https://github.com/cruiseresearchgroup/SOCIA/tree/evo}.
\end{abstract}

\section{Introduction}
\label{sec:intro}
\begin{figure*}[t]
\centering
\includegraphics[width=1.0\textwidth]{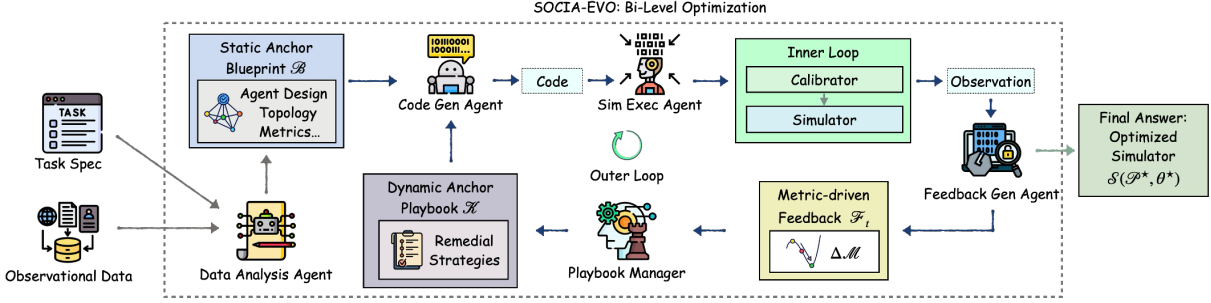} 
\caption{The \socia framework. The process is dual-anchored by a static \textit{Blueprint} ($\mathcal{B}$) and a dynamic \textit{Strategy Playbook} ($\mathcal{K}$). A bi-level optimization decouples structural refinement (Outer Loop) from parameter calibration (Inner Loop), leveraging metric-driven feedback to evolve strategies and prevent optimization instability.}
\label{fig:framework}
\end{figure*}

The automated construction of simulators from observational data—\emph{data-driven simulation}—is a cornerstone for understanding complex systems~\cite{brunton2016discovering,venkatramanan2018using,lejarza2022data,monti2023learning}. Unlike generic software engineering where functional correctness suffices, simulator construction is fundamentally a \emph{scientific modeling} task requiring \emph{distributional fidelity}~\cite{cranmer2020frontier,park2023generative,argyle2023out}. The generated program must reproduce the statistical regularities, causal mechanisms, and emergent behaviors of the ground truth. While Large Language Models (LLMs) demonstrate strong capabilities in translating natural language into code~\cite{chen2021evaluating,li2022competition,fried2022incoder}, applying them to this domain remains challenging~\cite{jimenez2023swe,liu2023your,la2024code}. Moving from a static dataset to a dynamic, executable simulation requires bridging a semantic gap while maintaining rigorous consistency with empirical constraints~\cite{epstein1999agent,camargo2020automated,gao2023s3,wang2024survey}.

\vspace{0.25em}
\noindent
We identify two critical failure modes that arise when standard LLM-based agents are applied to automated simulator construction over long horizons.
The first is \emph{contextual drift}.
As simulator complexity increases---with intricate state transitions, heterogeneous entities, and rich interaction dynamics---constraints established during the initial data analysis gradually lose salience in the model's attention~\cite{liu2024lost}.
Agents may begin to hallucinate governing rules, violate data schemas, or introduce mechanisms that were never supported by evidence~\cite{DBLP:conf/aaai/TianYYZCWLMS25,DBLP:journals/corr/abs-2504-20799}.
Without a persistent and authoritative anchor, the agent's implicit objective can shift from \emph{data fidelity} to mere \emph{code completion}~\cite{Arike_Donoway_Bartsch_Hobbhahn_2025}, yielding simulators that are syntactically plausible yet physically or statistically invalid~\cite{lee2024wrote,xi2025rise,DBLP:journals/corr/abs-2503-15223}.

\vspace{0.25em}
\noindent
The second, and more critical, failure mode is \emph{optimization instability} during calibration. 
First, agents frequently \emph{conflate structure and parameters}, failing to distinguish structural flaws (e.g., incorrect transition logic) from parametric misalignment (e.g., suboptimal rates). 
An agent may rewrite otherwise-correct logic when simple tuning would suffice, leading to oscillatory modifications analogous to a `whack-a-mole' game~\cite{mcculloch2022calibrating,olausson2023self,huang2025memorb}.
Second, this instability is amplified by the absence of a persistent mechanism for \emph{actionable remediation}~\cite{madaan2023self}. As the context window advances, the agent loses access to previously attempted fixes and their quantitative outcomes, and thus repeatedly proposes superficially plausible but empirically ineffective repairs~\cite{tan2025prospect,zhang2025survey}.
Crucially, these repairs are not ``truth'' but \emph{hypotheses}: an LLM may misattribute errors or suggest misleading fixes~\cite{huang2023large,liu2024exploring,zhang2025llm}. 
Without a system to \emph{validate} and \emph{refine} such hypotheses through repeated trials, the agent accumulates brittle advice, revisits failures, and struggles to converge to a high-fidelity simulator~\cite{wang2023voyager,mundler2024swt}.

\vspace{0.25em}
\noindent
Recent work tackles pieces of long-horizon LLM agents, but lacks a holistic solution for \emph{data-driven simulator construction}. General context-evolution methods—Reflexion~\cite{shinn2023reflexion}, TextGrad~\cite{DBLP:journals/corr/abs-2406-07496}, Dynamic Cheatsheets~\cite{suzgun2025dynamic}, and ACE~\cite{zhang2025agentic}—do not provide \emph{metric-grounded remediation validation}: they may use execution feedback, yet lack persistent attribution of improvements to specific hypotheses and principled falsification of ineffective ones. In contrast, LLM simulation frameworks like YuLan-OneSim~\cite{wang2025yulan} generate behaviors \emph{within} pre-defined environments, assuming simulator rules/topology are fixed rather than \emph{constructed} from observational data. Even calibration-aware methods such as G-Sim~\cite{DBLP:journals/corr/abs-2506-09272} are typically short-horizon optimizers without persistent tracking and validation of remedial strategies through repeated execution and metric gains, leading to re-exploration of ineffective fixes and optimization instability.

\vspace{0.25em}
\noindent
\textbf{\socia} (\textbf{S}imulation \textbf{O}rchestration for \textbf{C}omputational \textbf{I}ntelligence with \textbf{A}gents-\textbf{Evo}lutionary) reconceptualizes automated simulator construction as a \emph{dual-anchored evolutionary} process. \textbf{First}, to mitigate contextual drift, we establish a \emph{Blueprint}---a static, authoritative specification derived from data analysis that anchors the agent to immutable statistical constraints. \textbf{Second}, we implement a \emph{bi-level optimization} strategy. By delegating continuous parameter tuning to an inner numerical loop, we decouple structural reasoning from parameter search, ensuring that agent feedback targets intrinsic structural limitations rather than parameter noise. \textbf{Third}, we maintain a \emph{Playbook} of \emph{Remedial Strategies}: candidate repair hypotheses distilled from metric-driven diagnoses. Importantly, strategies are not assumed correct; they are \emph{validated in action}. 
Strategies that repeatedly improve metrics are promoted, while ineffective or harmful hypotheses are \emph{empirically falsified} and down-weighted. This enables the Playbook to \emph{self-curate}, retaining only high-utility strategies to prevent regressions and redundant exploration.

\vspace{0.25em}
\noindent
\textbf{Contributions.}
\textbf{(1)} We formalize \emph{simulator construction} as a dual-anchored agentic task, distinguishing it from generic code generation by emphasizing \emph{distributional fidelity} and \emph{optimization instability}.
\textbf{(2)} We propose \socia, introducing (i) a bi-level optimization architecture to decouple parameter calibration from structural refinement, and (ii) a \emph{Blueprint} to enforce consistency with task constraints.
\textbf{(3)} We design a \emph{metric-driven Playbook} that maintains \emph{Remedial Strategies} via a state transition system and a knapsack-based retrieval policy. Strategies are \emph{validated by execution}: those that consistently yield improvements are retained with higher reliability, while \emph{falsified strategies} are down-weighted, enabling self-curation and stabilizing long-horizon optimization.

\section{Related Work}
\label{sec:related_work}

\socia addresses the inverse problem of \emph{inferring} simulator logic from data, unlike Generative Agents~\cite{park2023generative} and YuLan-OneSim~\cite{wang2025yulan} which operate \emph{within} fixed environments. This objective challenges general refinement (e.g., Reflexion~\cite{shinn2023reflexion}, TextGrad~\cite{DBLP:journals/corr/abs-2406-07496}) and APR methods~\cite{DBLP:conf/icse/XiaWZ23,DBLP:conf/iclr/ChenLSZ24,DBLP:conf/icse/BouzeniaDP25}, whose reliance on functional correctness or deterministic oracles is incompatible with stochastic simulation, often leading to misattributed logic faults. Even calibration-aware approaches like G-Sim~\cite{DBLP:journals/corr/abs-2506-09272} risk instability by conflating structural and parametric errors. Finally, distinct from accumulative-context frameworks like MemGPT~\cite{packer2023memgpt} or ACE~\cite{zhang2025agentic}, \socia employs a self-curating Strategy Playbook to \emph{falsify} ineffective hypotheses, mitigating hallucinations over long horizons.

\section{The \socia Framework}
\label{sec:methodology}
\subsection{Problem Formulation \& Overview}
\label{sec:problem_formulation}

We formalize Automated Simulator Construction as a search problem within the space of executable programs. 
As shown in Figure~\ref{fig:framework}, given an observational dataset $\mathcal{D}_{obs}$ and a task specification $\mathcal{I}$, the goal is to synthesize a simulator $S$ parameterized by discrete code structure $P$ and continuous parameters $\theta$. 
Let $\mathcal{D}_{sim} \sim S(P, \theta)$ denote the simulated dataset obtained by rolling out the simulator (with stochasticity induced by simulator randomness and/or Monte Carlo sampling). 
The optimization objective is to minimize the distributional distance between $\mathcal{D}_{sim}$ and $\mathcal{D}_{obs}$:
\begin{equation}
    (P^*, \theta^*) = \mathop{\arg\min}_{P, \theta} \text{Dist}\left( \mathcal{D}_{sim}, \mathcal{D}_{obs} \right)
\end{equation}
, where $\mathcal{D}_{sim} \sim S(P, \theta)$.
To solve this problem efficiently, \socia employs a closed-loop iterative workflow involving six specialized agents. 
As outlined in Algorithm~\ref{alg:socia_ace}, the process begins with: (1) \textbf{Data Analysis Agent}, which synthesizes a static \textit{Blueprint} ($\mathcal{B}$) to anchor the search space. 
The system then enters an evolutionary loop: 
(2) The \textbf{Code Generation Agent} synthesizes simulator code $P_t$ and a parameter calibrator $C_t$ based on $\mathcal{B}$ and the current \textit{Playbook} strategies $\mathcal{K}$; 
(3) The \textbf{Simulation Execution Agent} executes $C_t$ to optimize $\theta$ (inner loop) and runs the simulation to obtain metrics $\mathcal{M}_t$; 
(4) The \textbf{Feedback Generation Agent} diagnoses discrepancies between $\mathcal{M}_t$ and $\mathcal{B}$; 
(5) The \textbf{Playbook Manager} updates the remedial strategy repository $\mathcal{K}$; 
(6) The \textbf{Iteration Control Agent} evaluates convergence to decide termination.

\begin{algorithm}[t]
\caption{\socia Framework Flow}
\label{alg:socia_ace}
\begin{algorithmic}[1]
\Require Observational Data $\mathcal{D}_{obs}$, User Intent $\mathcal{I}$
\Ensure Optimized Simulator $S(P^*, \theta^*)$
\State $\mathcal{B} \gets \text{DataAnalysisAgent}(\mathcal{D}_{obs}, \mathcal{I})$ \Comment{Generate Blueprint (Static Anchor)}
\State $\mathcal{K} \gets \emptyset$, $P_{-1} \gets \emptyset$ \Comment{Initialize Playbook and Code History}
\State $t \gets 0$
\While{not \text{Converged}}
    \State $\mathcal{S}_{sel} \gets \text{Knapsack}(\mathcal{K})$ \Comment{Select strategies from Playbook}
    
    \State $P_t, C_t \gets \text{CodeGenAgent}(\mathcal{B}, \mathcal{S}_{sel}, P_{t-1})$
    
    \State $\theta_t^* \gets \text{CalibratorExec}(C_t, \mathcal{D}_{obs})$ \Comment{Optimize continuous params}
    \State $\mathcal{M}_t \gets \text{SimExecAgent}(P_t, \theta_t^*)$ \Comment{Get metrics (Distribution fidelity)}
    
    \State $F_t \gets \text{FeedbackGenAgent}(\mathcal{M}_t, \mathcal{B}, P_t)$
    
    \State $\mathcal{K} \gets \text{UpdatePlaybook}(\mathcal{K}, F_t)$ \Comment{Merge \& State Update (Open/Resolved)}
    
    \If{$\text{IterationControlAgent}(\frac{\Delta \mathcal{M}}{\mathcal{M}_t})$ is \textsc{Stop}}
        \State \textbf{break}
    \EndIf
    \State $t \gets t + 1$
\EndWhile
\State \textbf{return} $S(P_t, \theta_t^*)$
\end{algorithmic}
\end{algorithm}

\subsection{The Dual-Anchoring Mechanism}
\label{sec:anchors}

To mitigate context drift and ensure consistent long-horizon optimization, \socia establishes two complementary memory structures: a static \textit{Blueprint} that enforces immutable constraints, and a dynamic \textit{Strategy Playbook} that evolves with the optimization trajectory.

\subsubsection{The Static Anchor: Blueprint}
\label{sssec:anchor_blueprint}
The \textbf{Blueprint} ($\mathcal{B}$) serves as the authoritative specification for the simulator. Unlike dynamic prompts or file summaries that evolve and potentially decay over long horizons, $\mathcal{B}$ functions as a set of \textbf{immutable constraints} anchoring the generative process to the ground truth. 
Structurally, $\mathcal{B}$ rigorously defines the simulation topology, agent schemas (roles, states, attributes), exogenous signals, and crucial evaluation metrics aligned with $\mathcal{D}_{obs}$. 
To ensure domain alignment, we incorporate a human-in-the-loop verification step where experts review and iteratively refine the initial $\mathcal{B}$ via natural language feedback before the optimization loop begins. 
This preemptive intervention prevents the search from initializing in an invalid subspace, ensuring that all subsequent iterations adhere to a valid trajectory verified by domain knowledge.

\subsubsection{The Dynamic Anchor: Playbook}
While the Blueprint fixes the target, the path to reach it requires adaptive memory. The \textbf{Strategy Playbook} ($\mathcal{K}$) acts as a dynamic repository of \emph{Remedial Strategies}---actionable repair hypotheses. Unlike passive logs, the Playbook is structured to facilitate evidence-based retrieval and self-curation.
We formally define the Playbook as a set of strategies $\mathcal{K} = \{S_1, S_2, \dots, S_N\}$. Each strategy $S_i$ is a structured tuple $S_i = \langle R_i, I_i, \Sigma_i \rangle$:
(1) \textbf{Reflection ($R_i$)}: Encapsulates the diagnostic content, including the identified root cause, the proposed corrective approach, and critically, the \textit{metric links} set $\Lambda_i \subset \mathcal{M}$. The set $\Lambda_i$ explicitly binds the structural defect to specific simulation metrics (defined in $\mathcal{B}$), enabling evidence-based retrieval.
(2) \textbf{Meta-info ($I_i$)}: Maintains evolutionary statistics to estimate reliability and scheduling priority. 
It is a quadruple $(u_i, un_i, s_i, f_i)$, where $u_i$ is the \textit{usage count} (times selected into the prompt), 
$un_i$ is the \textit{unusage count} (times not selected), 
$s_i$ is the \textit{success attribution} (times linked metrics improve and the issue is resolved), and 
$f_i$ is the \textit{failure attribution} (times linked metrics regress after selection), respectively.
(3) \textbf{State}: A label from the finite state set indicating the strategy's current lifecycle phase.

\subsection{The Core Engine: Bi-Level Optimization}
\label{sec:bilevel}

With the anchors established, we now define the execution engine. A core challenge in simulator construction is that structural flaws (e.g., missing feedback loops) and parametric misalignment (e.g., incorrect coefficients) often manifest as similar metric deviations. 
LLMs, while adept at discrete logical reasoning, struggle with high-dimensional continuous parameter search. 
To address this, we formulate the code generation process as a \textbf{bi-level optimization problem}, decoupling the discrete search for program structure $P$ from the continuous optimization of parameters $\theta$.

\paragraph{Outer Loop: Structural Refinement.}
The outer loop performs a discrete search over the space of valid programs. At iteration $t$, the Code Generation Agent acts as a policy $\pi_{code}$, synthesizing the simulator structure $P_t$ and a corresponding parameter calibrator $C_t$:
\begin{equation}
    (P_t, C_t) \leftarrow \pi_{code}(P_{t-1}, \mathcal{B}, \text{Knapsack}(\mathcal{K}))
\end{equation}
where $\mathcal{B}$ ensures schema compliance and $\mathcal{K}$ provides evolutionary strategies. Crucially, the agent does not guess $\theta$; instead, it synthesizes the \textit{optimization procedure} ($C_t$) required to find them. Specifically, the agent translates the parameter constraints (ranges, types) specified in $\mathcal{B}$ into executable search spaces (e.g., Optuna distributions) within $C_t$, ensuring the optimization is strictly bounded by domain knowledge.

\paragraph{Inner Loop: Parameter Calibration.}
The inner loop executes the synthesized calibrator $C_t$, which employs a numerical optimizer (e.g., Bayesian optimization or random calibrator) to solve for the optimal parameters $\theta_t^*$ under the fixed structure $P_t$:
\begin{equation}
    \theta_t^* = \mathop{\arg\min}_{\theta \in \Theta} \mathcal{L}\left( \text{Sim}(P_t, \theta), \mathcal{D}_{obs} \right)
\end{equation}
where $\mathcal{L}$ is a dynamically constructed loss function that aggregates the distance metrics defined in $\mathcal{B}$, and $\Theta$ is the feasible parameter domain.
This ensures that the performance metrics observed by the Feedback Agent represent the \textit{intrinsic capacity} of the structure $P_t$. 
By evaluating $P_t$ at its parametric optimum, the system effectively filters out noise from untuned parameters, allowing the Feedback Agent to accurately attribute remaining deficits to structural logic rather than calibration errors.

\paragraph{Iteration Control Policy.}
Iteration control stops when improvements plateau or regress, preventing over-correction and saving compute. 

\subsection{The Evolutionary Loop: Diagnosis, Curation \& Retrieval}
\label{sec:evolutionary_loop}

The core engine generates a simulator and metrics, but convergence requires a mechanism to learn from errors over time. We close the loop via a three-stage evolutionary process: (1) Evidence-based diagnosis, (2) Strategy lifecycle management, and (3) Context engineering for the next iteration.

\subsubsection{Stage 1: Evidence-Based Diagnosis}
To bridge numerical simulation results with symbolic reasoning, the Feedback Agent performs \textbf{Evidence-Based Diagnosis} within a strictly constrained context. 
The agent is restricted to an authoritative whitelist of metric keys ($\mathbb{K}_{metric}$) derived from the Blueprint, preventing the hallucination of non-existent evaluation criteria.

We enforce \textbf{schema-constrained attribution}: every identified defect must explicitly bind to specific metrics via a \texttt{metric\_links} field, validated by exact string matching against $\mathbb{K}_{metric}$. This filters out generic qualitative complaints and enforces a rigorous mapping from defects to quantitative evidence.
Furthermore, to ensure actionable repairs, we employ \textbf{structured diagnosis fields}. The agent must populate \textbf{four} distinct abstraction levels: 
(1) \textbf{Symptom Translation} (converting numeric signals into natural language); 
(2) \textbf{Mechanism Hypothesis} (pinpointing responsible code logic and line numbers); 
(3) \textbf{Remediation Strategy} (formulating high-level directives for the next iteration); 
and (4) \textbf{Severity Assessment} (classifying the defect criticality into discrete levels, i.e., \texttt{BLOCKER}, \texttt{HIGH}, \texttt{MEDIUM}, \texttt{LOW}, to prioritize repair urgency). 
Only feedback satisfying this structural integrity is accepted into the Playbook.

\subsubsection{Stage 2: Lifecycle Management \& Self-Curation}
\label{ssec:lifecycle}
Once diagnosed, strategies enter the Playbook where their utility is tested over time. We model this as a transition system over $\mathcal{Q}=\{\textsc{Open}, \textsc{Queued}, \textsc{InProgress}, \textsc{Resolved}\}$, with full transition rules summarized in Table~\ref{tab:state_transition}.

\paragraph{Phase 1: Selection \& Lifecycle Events.}
These events govern strategy admission into the context window and backlog management.
$E_{new}$ initializes a new strategy $S_{new}$ in \textsc{Open} when no match exists in $\mathcal{K}$; 
$E_{merge}$ refreshes a matched historical strategy $S_{hist}$ (resetting it to \textsc{Open} while retaining its counters) when the same issue recurs; 
$E_{selected}$/$not\_selected$ are triggered by knapsack scheduling (Stage~3), moving selected strategies to \textsc{InProgress} with $u_i \leftarrow u_i+1$, and routing unselected ones to (or keeping them in) \textsc{Queued} with $un_i \leftarrow un_i+1$ for backlog aging.

\paragraph{Phase 2: Metric-Driven Feedback Events.}
We refer $\frac{\Delta \mathcal{M}}{\mathcal{M}_t}$ to the average relative change of the linked metric set $\Lambda_i$ from iteration $t-1$ to $t$, stabilized by a small $\epsilon$ to avoid division by zero.
\begin{equation}
\frac{\Delta \mathcal{M}}{\mathcal{M}_t}
:= \frac{1}{|\Lambda_i|}\sum_{m\in \Lambda_i}\frac{M_t(m)-M_{t-1}(m)}{|M_{t-1}(m)|+\epsilon}.
\end{equation}
Upon execution of the bi-level optimization, the resulting metric changes ($\frac{\Delta \mathcal{M}}{\mathcal{M}_t}$) trigger evaluation events for strategies in \textsc{InProgress}:
\textbf{(1) $E_{resolved}$ (Validation Success)}: Triggered when linked metrics improve significantly ($\frac{\Delta \mathcal{M}}{\mathcal{M}_t} > \tau$\footnote{We set $\tau$ as 3\% empirically.}). The strategy transitions to \textsc{Resolved}, and the success counter is incremented ($s_i \leftarrow s_i + 1$). This provides \emph{empirical validation} that the remediation hypothesis is beneficial.
\textbf{(2) $E_{falsified}$ (Falsification)}: Triggered when linked metrics degrade significantly ($\frac{\Delta \mathcal{M}}{\mathcal{M}_t} < -\tau$). This suggests the hypothesis was ineffective or harmful. The strategy returns to \textsc{Open} for reconsideration, while its failure counter is incremented ($f_i \leftarrow f_i + 1$). Accumulating failures reduces future selection probability, enabling \emph{self-curation}.
\textbf{(3) $E_{uncertain}$ (Inconclusive)}: Triggered when metric changes are negligible ($|\frac{\Delta \mathcal{M}}{\mathcal{M}_t}| \le \tau$) or inconsistent. The strategy returns to \textsc{Open} with $u_i \leftarrow u_i + 1$, avoiding premature confirmation or rejection.
We merge semantically similar feedback into an existing strategy (thresholded matching) and inherit its $(u_i,{un}_i,s_i,f_i)$ to preserve an evidence-based reliability estimate.

\begin{table}[t]
\centering
\small
\caption{State Transition Logic for Playbook.}
\label{tab:state_transition}
\begin{tabular}{@{}llc@{}}
\toprule
\textbf{Current State} & \textbf{Event} & \textbf{New State} \\ \midrule
(None) & $E_{new}$ & \textsc{Open}  \\
\textsc{Open} & $E_{selected}$ & \textsc{InProgress} \\
\textsc{Queued} & $E_{selected}$ & \textsc{InProgress}  \\
\textsc{Open} & $not\_selected$ & \textsc{Queued}  \\
\textsc{Queued} & $not\_selected$ & \textsc{Queued}  \\ \midrule
\textsc{InProgress} & $E_{resolved}$ & \textsc{Resolved}  \\
\textsc{InProgress} & $E_{falsified}$ & \textsc{Open}  \\
\textsc{InProgress} & $E_{uncertain}$ & \textsc{Open} \\ \midrule
$\forall \Sigma \in \mathcal{Q}$ & $E_{merge}$ & \textsc{Open}  \\ \bottomrule
\end{tabular}
\end{table}

\subsubsection{Stage 3: Context Engineering via Knapsack Selection}
\label{ssec:knapsack}
To construct the next-iteration prompt under a limited context budget $L_{budget}$, we select a subset of candidate strategies $\mathcal{S}_{sel}$ from the \textsc{Open}/\textsc{Queued} pool by solving a standard 0--1 knapsack:
\begin{equation}
\max_{\mathbf{x}\in\{0,1\}^{|\mathcal{K}_{cand}|}} \sum_{i} v_i x_i
\quad \text{s.t.} \quad
\sum_{i} c_i x_i \le L_{budget},
\end{equation}
where $c_i$ is the token cost of including strategy $S_i$ and $v_i$ is its estimated utility.

\paragraph{Valuation Function.}
We define the utility as a product of severity, backlog urgency, and reliability:
\begin{equation}
    v_i = w_{sev}\cdot U_i^{queue}\cdot \Phi_{rel}(S_i),
\end{equation}
where $w_{sev}$ is a discrete severity weight mapped from the Severity Assessment:
$w_{sev}(\textsc{Blocker})=1.0$, $w_{sev}(\textsc{High})=0.8$, $w_{sev}(\textsc{Medium})=0.4$, and $w_{sev}(\textsc{Low})=0.2$. 
We use this monotone mapping to prioritize high-impact defects while keeping the scale bounded.
 $U_i^{queue}=1+\lambda \cdot \min(un_i, K_q)$ is a backlog bonus that prevents starvation: $un_i$ counts how many times $S_i$ was not selected by the knapsack scheduler, capped by $K_q$.
We set $\lambda=0.05$ and $K_q=10$ in experiments.
$\Phi_{rel}$ is the empirical reliability, and we compute it as follows.


\paragraph{Bayesian Self-Curation.}
We model $\Phi_{rel}(S_i)$ as the posterior mean of a Beta--Bernoulli model with a uniform prior:
\begin{equation}
    \Phi_{rel}(S_i)=\mathbb{E}[\mathrm{Beta}(s_i+1,f_i+1)]
    =\frac{s_i+1}{s_i+f_i+2},
\end{equation}
where $s_i$ and $f_i$ denote the accumulated success and failure attributions of strategy $S_i$, respectively. 
This formulation explicitly treats each synthesized strategy as a \emph{hypothesis} rather than a trusted correction. In other words, we do not assume that a strategy generated by the LLM is correct a priori. Instead, its utility is estimated only through execution-grounded evidence: if the strategy is selected and repeatedly leads to improvements on its linked metrics, its success count $s_i$ increases, which raises $\Phi_{rel}(S_i)$ and therefore its future selection priority; conversely, if applying the strategy leads to regressions or consistently fails to resolve the targeted issue, its failure count $f_i$ increases, which monotonically lowers $\Phi_{rel}(S_i)$ and suppresses it in subsequent retrieval.

\paragraph{Strategic Prompt Layout.}
To counter the ``Lost in the Middle'' effect~\cite{liu2024lost,hsieh2024found,guo2025serial}, we structure the prompt into three zones: (1) System Zone (Primacy) for agent's role definitions and task instructions; (2) Background Zone (Middle) for logs and previous code; and (3) \textbf{Instruction Zone (Recency)}, where the Blueprint $\mathcal{B}$ and the selected strategies $\mathcal{S}_{sel}$ are placed immediately preceding the generation token. This ensures the agent conditions its generation on the most critical constraints and validated fixes.

\section{Experimental Setup}
\label{sec:exp_settings}

\begin{table*}[t]
  \centering
  \small
  \caption{Evaluation results on three simulation tasks. Values are reported as mean $\pm$ 95\% CI, and lower values indicate better performance. The best and second-best results are highlighted in \textbf{bold} and \underline{underlined}, respectively. \textbf{Mob. N$\rightarrow$N}, \textbf{Mob. A$\rightarrow$A}, and \textbf{Mob. N$\rightarrow$A} denote \textbf{Personal Mobility} under normal-to-normal prediction, abnormal-to-abnormal (pandemic) prediction, and normal-to-abnormal generalization, respectively.}
  \label{tab:overall-results}
  \setlength{\tabcolsep}{5pt}
  \renewcommand{\arraystretch}{1.18}
  \resizebox{\textwidth}{!}{
  \begin{tabular}{lcccccccc}
    \toprule
    \multirow{2}{*}{\textbf{Methods} $\downarrow$} 
    & \textbf{User.} 
    & \textbf{Mask.} 
    & \multicolumn{2}{c}{\textbf{Mob. N$\rightarrow$N (ID)}} 
    & \multicolumn{2}{c}{\textbf{Mob. A$\rightarrow$A (ID)}} 
    & \multicolumn{2}{c}{\textbf{Mob. N$\rightarrow$A (OOD)}} \\
    \cmidrule(lr){2-2} \cmidrule(lr){3-3} \cmidrule(lr){4-5} \cmidrule(lr){6-7} \cmidrule(lr){8-9}
    & MAE $\downarrow$ 
    & RMSE $\downarrow$ 
    & JSD $\downarrow$ 
    & WD $\downarrow$ 
    & JSD $\downarrow$ 
    & WD $\downarrow$ 
    & JSD $\downarrow$ 
    & WD $\downarrow$ \\
    \midrule
    Reflexion    & 0.17$\pm$0.010 & 0.26$\pm$0.017 & 0.16$\pm$0.011 & 0.52$\pm$0.016 & 0.18$\pm$0.016 & 0.53$\pm$0.016 & 0.16$\pm$0.017 & 0.69$\pm$0.020 \\
    Yulan-OneSim & 0.21$\pm$0.018 & 0.16$\pm$0.014 & 0.12$\pm$0.016 & 0.49$\pm$0.017 & 0.16$\pm$0.014 & 0.51$\pm$0.015 & 0.13$\pm$0.017 & 0.64$\pm$0.014 \\
    G-SIM-ES     & \underline{0.13$\pm$0.013} & 0.27$\pm$0.018 & \underline{0.07$\pm$0.010} & \underline{0.38$\pm$0.018} & \underline{0.07$\pm$0.014} & 0.46$\pm$0.014 & 0.18$\pm$0.015 & 0.64$\pm$0.012 \\
    G-SIM-SBI    & 0.19$\pm$0.014 & \underline{0.11$\pm$0.019} & 0.14$\pm$0.012 & 0.42$\pm$0.013 & 0.08$\pm$0.013 & \underline{0.43$\pm$0.013} & \textbf{0.06$\pm$0.018} & \underline{0.56$\pm$0.015} \\
    DC-CU        & 0.27$\pm$0.010 & 0.29$\pm$0.015 & 0.19$\pm$0.013 & 0.58$\pm$0.014 & 0.24$\pm$0.014 & 0.60$\pm$0.015 & 0.20$\pm$0.015 & 0.71$\pm$0.011 \\
    ACE-OL       & 0.14$\pm$0.015 & 0.24$\pm$0.014 & 0.10$\pm$0.015 & 0.44$\pm$0.017 & 0.14$\pm$0.017 & 0.50$\pm$0.015 & 0.11$\pm$0.014 & 0.61$\pm$0.017 \\
    \midrule
    \rowcolor{orange!15}
    \textbf{\socia} 
                 & \textbf{0.11$\pm$0.012} 
                 & \textbf{0.07$\pm$0.010} 
                 & \textbf{0.04$\pm$0.013} 
                 & \textbf{0.34$\pm$0.016} 
                 & \textbf{0.03$\pm$0.013} 
                 & \textbf{0.36$\pm$0.014} 
                 & \textbf{0.06$\pm$0.015} 
                 & \textbf{0.53$\pm$0.016} \\
    \bottomrule
  \end{tabular}
  }
\end{table*}

\subsection{Benchmarks} 
We evaluate \socia on three real-world–inspired simulation tasks covering distinct modeling challenges.
Unless otherwise specified, all datasets are in \textbf{English}, and the included human profile statistics represent general populations without targeting \emph{specific demographic groups}.
\textbf{(1) User Modeling.} Adapted from the AgentSociety Challenge~\cite{DBLP:journals/corr/abs-2502-18754}, this task requires simulating user behaviors to predict product ratings based on historical interactions.
The benchmark utilizes a rich corpus of \textbf{20,000 user-item reviews} for simulator calibration; we confirm that this public dataset was \emph{de-identified upon release}. The evaluation targets the accurate prediction of \textbf{1,200 specific user-item ratings}.

\textbf{(2) Mask Adoption Simulation.} Inspired by BESSIE~\cite{DBLP:journals/corr/abs-2203-11414} and pandemic decision models~\cite{mitsopoulos2023masking}, this task models the diffusion of mask-wearing behaviors in a socially embedded population of \textbf{100 residents}.
As this dataset is \emph{synthetically generated} for this study, it contains no PII.
It challenges the agent to recover causal mechanisms involving heterogeneous social ties and external interventions. The temporal split uses the \textbf{first 30 days} of behavioral data for calibration, reserving the \textbf{subsequent 10 days} for simulation and prediction assessment.

\textbf{(3) Personal Mobility Generation.} Using the real-world LLMob dataset~\cite{wang2024large}, this task focuses on predicting next-day spatiotemporal trajectories.
The dataset, which was \emph{fully anonymized prior to publication}, tracks \textbf{69 residents} with an average observation period of \textbf{124.10 active days} per user.
The task evaluates robustness under distribution shifts by requiring the simulator to generate a full-day activity trajectory for a \textbf{randomly sampled target day}, generalizing from normal periods to pandemic-disrupted scenarios.

\subsection{Baseline Methods and Implementation} 
\label{ssec:baseline_impl}
\textbf{Baselines.} We compare \socia against two categories of methods, all initialized with the same \socia-derived data summaries. First, \textit{simulation-specific methods}: \textbf{YuLan-OneSim}~\cite{wang2025yulan} structures scenarios via ODD protocols and optimizes code through a verify–repair loop; the \textbf{G-SIM} family~\cite{DBLP:journals/corr/abs-2506-09272} (including \textbf{G-SIM-ES} and \textbf{G-SIM-SBI}) combines LLM generation with gradient-free calibration for robust extrapolation. Second, we adapt \textit{general agentic frameworks} by feeding metric-driven execution evidence as feedback to trigger memory updates: \textbf{Reflexion}~\cite{shinn2023reflexion} (verbal reinforcement via episodic memory), \textbf{Dynamic Cheatsheet (DC-CU)}~\cite{suzgun2025dynamic} (accumulating successful strategies), and \textbf{ACE-Online}~\cite{zhang2025agentic} (iteratively curating context playbooks).

\textbf{Evaluation Metrics.} We report means with 95\% confidence intervals across \textbf{five} random seeds~\cite{colas2018many}. 
Metrics are specified in the \emph{Blueprint} to align with task-specific goals: for \textbf{User Modeling}, we use Mean Absolute Error (\textbf{MAE}) on star ratings~\cite{DBLP:conf/naacl/WangJCYZCFLHY24}; for \textbf{Mask Adoption}, we use \textbf{RMSE} to assess the temporal alignment of adoption rates. For \textbf{Personal Mobility}, we evaluate distributional fidelity using \textbf{JSD} (Jensen-Shannon divergence of arrival times) and \textbf{WD} (Wasserstein distance of geographic trip lengths), where lower scores denote better results.

\textbf{Fairness and Implementation.} All experiments utilize \textbf{GPT-5.1}~\cite{openai2025gpt5}. To ensure strictly fair comparisons, we employ two reproduction strategies. (1) \textbf{Unified Skeleton for General Agents}: We implement ACE, DC, and Reflexion within a shared framework identical to \socia. We augment these agents with the same numerical calibrator, retrieval evidence, and LLM interfaces, isolating the \emph{memory update policy} as the sole variable. (2) \textbf{System-Level Alignment for Domain Baselines}: YuLan-OneSim and G-SIM operate via their native pipelines (e.g., G-SIM's closed-loop calibration) to preserve architectural integrity. We enforce strict alignment on external constraints, including identical data visibility (anti-leakage rules), compute budgets (max iterations), and evaluation metrics. All baselines run in fully automated modes without human intervention. The HITL step is specific to Blueprint initialization in \socia and its contribution is quantified in \S~\ref{ssec:ablation_study}.

\textbf{Human-in-the-loop Blueprint verification.}
Before iterative code generation begins, we apply a lightweight HITL verification step to the initial Blueprint $\mathcal{B}$, following \S\ref{sssec:anchor_blueprint}. Domain experts briefly review the automatically synthesized Blueprint and refine it through natural-language feedback to correct obvious specification drift and ensure that the initial schema, constraints, and core mechanisms are aligned with the task. This step is used only at initialization, rather than to supervise later iterations. To keep the pipeline scalable, the verified Blueprint and accumulated Playbook are reused across runs within the same task/domain, including many-seed settings. Although each seed still reruns the bi-level optimization process, this reuse avoids re-specifying task constraints, reduces repeated early-stage mistakes, and allows subsequent runs to proceed fully automatically and in parallel. In practice, the HITL cost is amortized over many runs and remains comparable to a brief specification review rather than substantial ongoing manual intervention.

\section{Evaluation Results}
\label{sec:evaluation}
\subsection{Main Results}
Table~\ref{tab:overall-results} summarizes the performance of \socia against all baselines across three tasks. \socia achieves consistent best performance.
These numerical improvements are statistically significant ($p<0.05$); we report means with 95\% confidence intervals across random seeds.

\textbf{Comparison with General Agentic Frameworks.}
It is important to reiterate that to ensure fairness, \textbf{Reflexion}, \textbf{DC}, and \textbf{ACE} were augmented with the same numerical calibrator as \socia. Despite this enhancement, they struggle in high-fidelity simulation tasks.
\textbf{Reflexion} and \textbf{DC-CU} show significant deficits in the \textbf{Mask Adoption} task (RMSE 0.26 and 0.29, respectively, compared to \socia's 0.07).
This empirical evidence suggests that equipping general agents with calibration tools is insufficient if the underlying \emph{memory update policy} cannot distinguish between structural errors and parametric miscalibration.
\textbf{Reflexion} tends to overfit to specific failure cases (e.g., fixing a syntax error but ignoring distribution shift), while \textbf{DC-CU} suffers from negative transfer—retrieving ``successful'' remedial strategies from disparate contexts (e.g., a wrong network topology) that bias the simulation logic.
\textbf{ACE-OL} performs relatively better (RMSE 0.24) due to its structured playbook, yet it still lags behind. This confirms that without the \emph{Blueprint} to constrain the search space and \emph{Bi-level Optimization} to decouple logic from parameters, general agents largely ``hallucinate'' plausible-looking but statistically invalid mechanisms.

\textbf{Comparison with Simulation-Specific Methods.}
\socia also surpasses domain-specific baselines, though the margins are tighter.
\textbf{YuLan-OneSim} shows decent structural correctness (RMSE 0.16 in Mask) thanks to its ODD protocol, but lacks the fine-grained parameter optimization required for precise trajectory fitting (trailing in Mobility metrics).
The \textbf{G-SIM family} is the strongest competitor. \textbf{G-SIM-ES} performs well in simpler tasks (User Modeling), while \textbf{G-SIM-SBI} excels in complex calibration (Mask RMSE 0.11).
However, \socia achieves superior consistency (e.g., lower WD across all mobility settings).
We attribute this to \socia's \emph{Strategy Playbook} and \emph{Knapsack} mechanism. While G-SIM effectively separates structural generation ($\lambda$) from parameter estimation ($\omega$), its iterative loop lacks long-term evidence tracking over ``remedial strategies''.
As a result, it may repeatedly revisit remediation hypotheses that have already been empirically falsified, leading to a ``whack-a-mole'' pattern.

\textbf{Robustness under Distribution Shifts.}
The \textbf{Mobility (N$\rightarrow$A)} task evaluates OOD robustness (training on normal, predicting pandemic).
\socia matches the OOD-specialized \textbf{G-SIM-SBI} in JSD (both \textbf{0.06}) but outperforms it in Wasserstein Distance (\textbf{0.53} vs. 0.56).
Since G-SIM-SBI is explicitly designed for posterior inference under uncertainty, its strong JSD is expected.
However, \socia's advantage in WD implies that our system captures the underlying \emph{spatial causal mechanism} (e.g., restricted mobility radius) more accurately than G-SIM, which likely relies more on fitting temporal distributions.
This verifies that \socia's \emph{Blueprint}-anchored reasoning prevents the model from merely overfitting parameters to the training distribution, enabling true mechanism recovery.


\begin{table}[t]
  \centering
  \small
  \caption{We conducted an ablation study using \socia as the baseline, where larger positive $\Delta$ values indicate greater performance degradation.}
  \label{tab:ablation_study}
  \setlength{\tabcolsep}{6pt}
  \renewcommand{\arraystretch}{1.18}
  \begin{tabular}{lccc}
    \toprule
    \multirow{2}{*}{\textbf{Models} $\downarrow$} & \textbf{User.} & \textbf{Mask.} & \textbf{Mob. N$\rightarrow$A} \\
    & $\Delta$MAE $\downarrow$ & $\Delta$RMSE $\downarrow$ & $\Delta$WD $\downarrow$ \\
    \midrule
    \rowcolor{orange!15}
    \textbf{SOCIA} & -- & -- & -- \\
    \midrule
    w/o \emph{inner}        & +0.25$\pm$0.01 & +0.47$\pm$0.01 & +0.29$\pm$0.01 \\
    w/o \emph{$\mathcal{B}$} & +0.20$\pm$0.02 & +0.37$\pm$0.01 & +0.23$\pm$0.01 \\
    w/o \emph{HITL}         & +0.18$\pm$0.02 & +0.34$\pm$0.02 & +0.21$\pm$0.02 \\
    w/o \emph{mem.}         & +0.14$\pm$0.01 & +0.30$\pm$0.02 & +0.20$\pm$0.01 \\
    w/o \emph{$\mathcal{K}$} & +0.10$\pm$0.02 & +0.23$\pm$0.02 & +0.15$\pm$0.02 \\
    w/o \emph{value}        & +0.08$\pm$0.01 & +0.17$\pm$0.01 & +0.10$\pm$0.02 \\
    \midrule
    -600   & +0.05$\pm$0.01 & +0.07$\pm$0.01 & +0.06$\pm$0.01 \\
    +600   & +0.02$\pm$0.02 & +0.04$\pm$0.03 & +0.03$\pm$0.02 \\
    +1200  & +0.09$\pm$0.01 & +0.11$\pm$0.01 & +0.09$\pm$0.01 \\
    +2200  & +0.12$\pm$0.01 & +0.14$\pm$0.02 & +0.13$\pm$0.01 \\
    \bottomrule
  \end{tabular}
\end{table}

\subsection{Ablation Study}
\label{ssec:ablation_study}
Table~\ref{tab:ablation_study} reports performance deltas relative to the full model, quantifying the contribution of each component in \socia.

\textbf{Component Impact.}
The largest drop comes from removing the inner numerical calibrator (\textbf{w/o \emph{inner}}): parameter updates degenerate into LLM-driven heuristic tuning from noisy metric feedback, which is coarse, unstable, and rarely yields consistent objective reduction, causing sharp fidelity loss (e.g., \textbf{+0.47 RMSE} in Mask Adoption) and underscoring \emph{bi-level decoupling}.
Removing the Blueprint (\textbf{w/o $\mathcal{B}$}) yields the second-largest degradation by increasing schema violations and weakening OOD robustness.
Ablating HITL (\textbf{w/o \emph{HITL}}) further hurts performance, indicating that lightweight expert checks during Blueprint construction prevent early mis-specifications from derailing the search.
Disabling the memory mechanism (\textbf{w/o \emph{mem.}}), where the agent operates without historical context, leads to optimization oscillations (``whack-a-mole'') observed in baselines.
Finally, replacing Knapsack retrieval with a sliding window (\textbf{w/o $\mathcal{K}$}) or removing value-based sorting (\textbf{w/o \emph{value}}) consistently impairs long-horizon convergence, as reliability-aware selection under a fixed context budget prevents low-value or redundant history from crowding out effective fixes.

\textbf{Context Window Analysis.}
We study sensitivity to the \emph{Recency Zone size} (default 1000 tokens; $\approx$5--10 strategies) under the `Lost in the Middle' effect. 
In Table~\ref{tab:ablation_study}, Shrinking it to 400 tokens (\textbf{-600}) hurts performance by dropping high-leverage strategies and guardrails.  
Conversely, expanding the zone yields diminishing returns (\textbf{+600}) or significant regression. 
At \textbf{+1200}, the overlong prompt triggers `Lost-in-the-Middle' effect.
At 3200 tokens (\textbf{+2200})—which allows fitting the entire Playbook—results in a sharp performance drop (e.g., +0.14 RMSE), approaching the degradation of removing memory.
This striking observation confirms that an overloaded context dilutes the model's attention as severely as having no memory at all.

\subsection{Analysis of Optimization Dynamics}
\label{ssec:opt_dynamics_summary}
To validate the stability and efficiency of \socia's iterative process, we conduct comprehensive supplementary experiments (detailed in \S~\ref{subsec:convergence_trajectory}, \S~\ref{app:cre_analysis}, and \S~\ref{ssec:resolution_dynamics}). 
First, by visualizing the \textbf{convergence trajectory}, we observe a distinct step-wise error reduction in early iterations followed by oscillatory trade-offs. This confirms the efficacy of our bi-level strategy in fixing global structural errors early on, while justifying our plateau-based termination policy to prevent over-correction during the later fine-tuning phase.

Second, to verify the suppression of catastrophic forgetting, we introduce a \textbf{Cumulative Recurrent Errors (CRE)} metric. Results show a \textbf{76\% reduction} in repeated mistakes by the transition into the fifth iteration, showing that our value-driven Knapsack mechanism automatically down-weights or retires empirically falsified strategies, thereby eliminating the ``whack-a-mole'' phenomenon observed in memory-less baselines.

Finally, analyzing the \textbf{Issue Resolution Rate (IRR)} reveals a transition from a ``rapid correction'' phase (resolving syntax/logic errors) to a ``stubborn bottleneck'' phase (tackling deep structural constraints). 
These analyses demonstrate that \socia does not stumble upon solutions; rather, it prunes the search space by down-weighting empirically falsified remedial strategies, while actively exploring and re-testing novel remediation hypotheses without regressing to known failures.

\subsection{Backbone Portability Analysis}
\label{sec:backbone-portability}

To examine whether \socia depends on a strong proprietary backbone, we replace GPT-5.1 in the agentic loop with two openly available instruction-tuned LLMs, \textit{Llama-3.3-70B-Instruct-Turbo} and \textit{Qwen3-Next-80B-A3B-Instruct}, while keeping the task setting, Blueprint construction, playbook mechanism, verification protocol, and iteration budget unchanged. We evaluate this setting on the \textbf{Mask Adoption} task from three complementary perspectives: task-level performance over the full iterative code-evolution process, framework-level diagnostics for recurrent-error suppression and issue-resolution effectiveness, and a qualitative analysis of the playbook entries produced by different backbones.

The results support three high-level conclusions. First, \socia is not intrinsically tied to a proprietary backbone: a sufficiently capable open-source model can achieve highly competitive, and sometimes superior task performance. Second, the core framework dynamics generalize across backbone choices, as open-source models preserve the same overall recurrent-error reduction pattern over iterations. Third, the remaining gap across backbones is mainly explained by differences in the quality of playbook construction and issue-to-repair execution, rather than by any inherent dependence of \socia on proprietary LLMs.
Detailed tables and representative examples are provided in \S~\ref{ssec:open-backbone}.

\subsection{Qualitative Code Comparison Across Optimization Iterations}
\label{sec:qualitative-code-comparison}


We compare the simulator code for the \textbf{Mobility task} across optimization iterations to assess whether \socia improves mainly through parameter retuning or structural refinement. We find that later iterations introduce concrete mechanism-level changes rather than merely adjusting calibrated coefficients, indicating that the outer loop translates numerical feedback into structural simulator refinement beyond inner-loop calibration. Details are provided in \S~\ref{append:qualitative_n2a}.

\section{Conclusion}
\label{sec:conclusion}
We presented \textbf{\socia}, a framework that treats simulator construction as a \emph{scientific modeling} discipline. 
To address contextual drift and optimization instability, \socia adopts a \emph{dual-anchored} architecture: a static \emph{Blueprint} to enforce empirical constraints, and a dynamic \emph{Strategy Playbook} that \emph{self-curates} remedial hypotheses via execution-grounded falsification. 
In addition, a \emph{bi-level optimization} decouples structural refinement from continuous parameter calibration, ensuring feedback targets intrinsic structural limitations rather than untuned parameter noise. 
Together, these mechanisms move LLM-based construction beyond semantic plausibility toward \emph{distributional fidelity}, with only lightweight, one-time human verification during Blueprint initialization.

\section*{Limitations}

\begin{enumerate}
\item \textbf{Backbone dependence.} Although the framework is model-agnostic at the architectural level, its practical performance remains bounded by the capability and bias of the underlying LLM. In highly novel, sparse, or underrepresented domains, the generated simulators may fail to recover semantically valid mechanisms or may overfit to patterns that are only weakly supported by data.

\item \textbf{Scope of supported dynamics.} The current instantiation focuses on tasks that can be expressed through iterative structural refinement and bounded parameter calibration. More complex settings involving long-horizon planning, strategic multi-agent interaction, or real-time adaptation under uncertainty may require stronger memory, richer reasoning modules, and tighter integration with causal and counterfactual modeling.

\item \textbf{Reproducibility and access.} Although open-source backbones can be competitive on some tasks (\S~\ref{sec:backbone-portability}), our strongest overall results rely on a commercial LLM backbone, which introduces dependency on external API access. Although the framework itself is not tied to a single proprietary model, this dependence may limit exact reproducibility and broad accessibility.
\end{enumerate}

\section*{Ethical Considerations}

\begin{enumerate}

\item \textbf{Ethics Approval and Oversight.}
The human feedback procedures in this study were reviewed and approved by the Research Ethics \& Compliance Support (RECS) at UNSW, and were conducted in accordance with the National Statement on Ethical Conduct in Human Research.

\item \textbf{Participant Recruitment and Consent.}
We recruited domain experts from social science and computing-related domains to assess and refine Blueprint fidelity and alignment. All participants received an information statement, provided written informed consent, and could withdraw at any time without penalty.

\item \textbf{Data Privacy and Management.}
All feedback data were de-identified prior to analysis. Personally identifiable information (PII) was removed, data were stored on secure institutional servers, and access was restricted to the research team.

\item \textbf{Potential Risks and Mitigation.}
Participation was assessed as low risk. Possible burdens were limited to minor frustration, reflection on professional experience, and time spent on evaluation tasks. To mitigate these risks, tasks were non-invasive, withdrawal was allowed at any time, and technical support was provided.

\item \textbf{Participant Instructions and Disclaimers.}
Participants were provided with detailed instructions and consent materials describing the workflow, data handling procedures, and relevant disclaimers. They were required to review these materials before participation.
Instruction materials are available in \S~\ref{sec:participant_instructions}.

\item \textbf{Recruitment Channels and Payment Adequacy.}
Recruitment was conducted through institutional mailing lists and professional networks associated with the authors' institution and partner universities. Participants were compensated with prepaid gift cards at a rate aligned with local minimum-wage and fair-work guidelines for research participation.

\item \textbf{Broader Impact.}
Although \socia aims to automate simulator construction, it is intended to augment rather than replace human expertise. The HITL verification step is included to reduce hallucination risk and keep generated simulators grounded in empirical constraints and ethical boundaries.

\end{enumerate}

\section*{Acknowledgments}
\label{acknowledgement}
This research is supported by the ARC Center of Excellence for Automated Decision Making and Society (CE200100005).
Computational facilities were provided by the School of Computer Science and Engineering at UNSW Sydney through the Wolfpack computational cluster. 
Additionally, we express our gratitude to the NVIDIA Academic Grant Program for providing access to the GPU resources on Saturn Cloud.
We used icon assets from \textit{Flaticon.com} to create the illustrations in this paper. We appreciate the artists who created and shared these icons on \url{Flaticon.com}.
We also thank Yang Liu (yang.liu39@student.unsw.edu.au ) for valuable discussions and contributions to the ideas behind this work.

\bibliography{custom}

\clearpage
\newpage
\appendix
\section{Supplementary Experiments}
\label{appen:add_exps}

\subsection{Experimental Setup and Resource Analysis}
\label{appendix:resource_consumption}

To ensure a fair comparison, we unified the experimental configurations for both the reproduction of all baselines and the evaluation of \socia.

\paragraph{Model and Inference Settings.} 
First, all agent implementations utilize the \textbf{GPT-5.1} version as the uniform backbone model. We standardized the reasoning intensity across tasks: the \textit{code generation phase} is set to \textbf{Medium} reasoning strength, while all other reasoning tasks (e.g., diagnosis, feedback generation, and patching) are set to \textbf{Low}.
Second, we unified the iteration budget. The maximum number of iteration rounds is set to \textbf{9}. If convergence is not achieved by this limit, the system automatically terminates the iterative process. We do not impose restrictions on the token throughput for LLM calls during these sessions.

\paragraph{Time and Convergence Analysis.}
As illustrated in Figure~\ref{fig:metric_trends} (\S \ref{subsec:convergence_trajectory}), statistical analysis indicates that \socia typically achieves optimal performance around the \textbf{3rd to 4th iteration}. Following a period of performance oscillation (approximately 2 rounds), the optimization process generally concludes at the \textbf{6th round} (with a maximum upper bound of 9 rounds).

The duration of each iteration round is approximately \textbf{30--50 minutes}, detailed as follows:
\begin{itemize}
    \item \textbf{LLM Latency:} The average waiting time for LLM API responses is \textbf{25 minutes}. Specifically, code generation (under \textit{Medium} reasoning) accounts for 10--15 minutes, while auxiliary tasks such as diagnosis and patch application (under \textit{Low} reasoning) average 1--2 minutes.
    \item \textbf{Simulation Runtime:} The remaining time is dedicated to simulator execution. This duration varies based on the complexity of the generated simulator and the parameter calibration epochs automatically specified by the code agent. Under normal execution conditions (without runtime errors), the total time for simulation, optimization, and inference ranges from a minimum of \textbf{60 seconds} to a maximum of \textbf{1,500 seconds}.
\end{itemize}

\paragraph{Economic Cost.}
For a complete lifecycle of \socia code generation (spanning the typical 6--7 rounds of optimization), the total token throughput results in an economic cost of approximately \textbf{\$1.50 -- \$2.00 USD}.

\subsection{Convergence Trajectory}
\label{subsec:convergence_trajectory}

\pgfplotstableread{
iter user_mae mask_rmse nn_jsd nn_wd aa_jsd aa_wd na_jsd na_wd
0    0.233    0.134     0.087  0.492  0.026  3.174  0.059  6.574
1    0.200    0.109     0.091  0.592  0.031  2.554  0.036  1.031
2    0.165    0.097     0.039  0.378  0.051  1.662  0.058  1.134
3    0.138    0.073     0.038  0.341  0.028  0.364  0.105  0.999
4    0.108    0.080     0.206  0.395  0.026  0.759  0.056  0.525
5    0.113    0.075     0.333  0.417  0.024  0.601  0.053  1.290
6    0.109    0.077     0.304  0.434  0.026  0.614  0.179  2.731
}\datatable

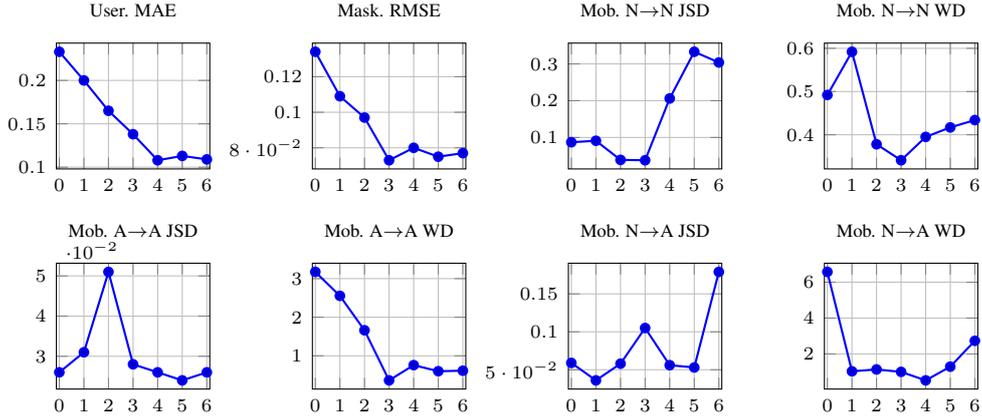
\begin{figure*}[t]
\centering
\begin{tikzpicture}

\begin{groupplot}[
  group style={
    group size=4 by 2,
    horizontal sep=1.35cm,
    vertical sep=1.25cm
  },
  width=0.225\textwidth,
  height=3.25cm,
  xmin=0, xmax=6,
  xtick={0,1,2,3,4,5,6},
  xlabel={},
  ylabel={},
  grid=both,
  tick label style={font=\scriptsize},
  label style={font=\scriptsize},
  title style={font=\scriptsize},
  enlarge x limits=0.02,
  enlarge y limits=0.08,
  every axis plot/.append style={
    mark=*,
    mark size=1.6pt,
    line width=0.8pt
  },
]

\nextgroupplot[title={User. MAE}]
\addplot table[x=iter,y=user_mae] {\datatable};

\nextgroupplot[title={Mask. RMSE}]
\addplot table[x=iter,y=mask_rmse] {\datatable};

\nextgroupplot[title={Mob. N$\to$N JSD}]
\addplot table[x=iter,y=nn_jsd] {\datatable};

\nextgroupplot[title={Mob. N$\to$N WD}]
\addplot table[x=iter,y=nn_wd] {\datatable};

\nextgroupplot[title={Mob. A$\to$A JSD}]
\addplot table[x=iter,y=aa_jsd] {\datatable};

\nextgroupplot[title={Mob. A$\to$A WD}]
\addplot table[x=iter,y=aa_wd] {\datatable};

\nextgroupplot[title={Mob. N$\to$A JSD}]
\addplot table[x=iter,y=na_jsd] {\datatable};

\nextgroupplot[title={Mob. N$\to$A WD}]
\addplot table[x=iter,y=na_wd] {\datatable};

\end{groupplot}



\end{tikzpicture}

\caption{Metric trends over iterations (small multiples). Each point is one iteration.}
\label{fig:metric_trends}
\end{figure*}

To understand how \socia improves simulator fidelity over iterative self-revision, we visualize its \emph{convergence trajectory} across three tasks and multiple metrics (as shown in Figure~\ref{fig:metric_trends}). The \textbf{x-axis} denotes iteration steps, and the \textbf{y-axis} reports \textbf{simulation error} (distance to ground truth; lower is better). We track eight curves, covering \textbf{User rating prediction} (MAE), \textbf{Mask adoption} (RMSE), and \textbf{Mobility simulation} under three regimes: \textbf{N$\rightarrow$N} (in-distribution), \textbf{A$\rightarrow$A} (in-distribution within abnormal period), and \textbf{N$\rightarrow$A} (OOD), each measured by \textbf{JSD} and \textbf{Wasserstein distance (WD)}.

Overall, \socia exhibits a \textbf{step-wise monotonic descent} in the early-to-mid iterations, followed by \textbf{oscillation/rebound} that triggers early stopping. Specifically, the first three iterations yield broad and consistent improvements: \emph{User MAE} drops from $0.233$ to $0.138$ (iter0$\rightarrow$iter3), and \emph{Mask RMSE} decreases from $0.134$ to $0.073$, indicating steadily improving predictive accuracy. For mobility, the gains are most pronounced in the WD-based errors: \emph{A$\rightarrow$A WD} sharply reduces from $3.174$ to $0.364$ by iter3, and \emph{N$\rightarrow$A WD} collapses from $6.574$ to $0.999$, suggesting the simulator quickly corrects coarse spatial displacement and trip-length mismatch. Meanwhile, \emph{N$\rightarrow$N JSD} improves from $0.087$ to $0.038$ (iter0$\rightarrow$iter3), showing better alignment of distributional structure under normal conditions.

After reaching this ``good'' region (around iter3--iter4), the curves begin to \textbf{oscillate}, revealing metric trade-offs and occasional over-correction. For instance, \emph{N$\rightarrow$N JSD} rebounds at iter4 ($0.206$) and further worsens at iter5 ($0.333$), while \emph{N$\rightarrow$A WD} improves at iter4 ($0.525$) but rebounds at iter5 ($1.290$) and degrades substantially at iter6 ($2.731$). These dynamics suggest that later edits can improve one regime while harming another, making further updates less reliably beneficial.

These trajectories support the design rationale of \ofsocia \textbf{bi-level strategy}: early iterations primarily fix high-impact, globally harmful errors (hence the staircase-like decreases), while later iterations become sensitive to metric trade-offs, where further edits can cause non-monotonic rebounds. Importantly, once such oscillation appears (e.g., iter5--iter6 shows clear degradation in multiple mobility errors), the \textbf{iteration control agent} terminates the loop, preventing drift away from previously validated improvements. This behavior aligns with our goal of \textbf{robust, stable progress}: \socia improves in discrete, verifiable steps until additional updates no longer yield reliable gains, at which point the controller halts to preserve the best-found simulator state.

\subsection{Quantification of Negative Knowledge Retention}
\label{app:cre_analysis}
To rigorously evaluate the efficiency of the \emph{Knapsack} mechanism in suppressing repeated mistakes (the ``whack-a-mole'' phenomenon), we introduce the \textbf{Cumulative Recurrent Errors (CRE)} metric. This metric quantifies the volume of structural or parametric errors generated at iteration $t$ that are semantically identical to strategies already falsified in iterations $0$ to $t-1$.

\paragraph{Definition and Detection Procedure.}
We maintain a dynamic \textit{Failure History Registry} $\mathcal{H}_{fail}$ containing normalized representations of all failed code snippets and strategies from previous iterations. 
The detection pipeline proceeds as follows:
\begin{enumerate}
    \item \textbf{Normalization:} For every generated simulator code $P_t, C_t$ or strategy description $S_t$, we strip variable naming permutations and formatting noise to extract a \textit{semantic fingerprint} (e.g., using Abstract Syntax Tree normalization for code).
    \item \textbf{Matching:} We compare the current fingerprint against $\mathcal{H}_{fail}$. A \textbf{Recurrent Error} is flagged if the similarity score (calculated via SequenceMatcher) exceeds a threshold of $0.95$, indicating the agent is retrying a known failure.
    \item \textbf{Calculation:} The CRE value for iteration $t$ is the count of such flagged errors within that generation batch.
\end{enumerate}

\paragraph{Experimental Setup.}
We report the CRE statistics across all three benchmarks: \textbf{User Modeling}, \textbf{Mask Adoption}, and \textbf{Personal Mobility} (including \textit{Normal$\rightarrow$Normal}, \textit{Abnormal$\rightarrow$Abnormal}, and \textit{Normal$\rightarrow$Abnormal} settings). 
Consistent with our main experiments (Section~\ref{ssec:baseline_impl}), results are averaged across five random seeds to ensure statistical robustness.

\paragraph{Results and Analysis.}
Table~\ref{tab:cre_results} presents the evolution of CRE from Iteration 1 to 5.
\textbf{Trend Analysis:} At the initial stage (Iter 1), the agent exhibits a higher tendency to repeat errors (Average CRE $\approx 2.20$), as the \emph{Playbook} is still sparse. 
However, as iterations progress, we observe a consistent and significant downward trend across all tasks.By Iteration 5, the average CRE drops to \textbf{0.53}, a reduction of \textbf{76\%} compared to Iteration 1.
\textbf{Task-Specific Insight:} The \textbf{Mask Adoption} and \textbf{Personal Mobility (N$\rightarrow$A)} tasks initially show high recurrence (1.8 and 2.6) due to the complexity of finding correct causal mechanisms (e.g., intervention logic).
Crucially, the rapid decline in these complex tasks confirms that the \emph{Knapsack} algorithm successfully identifies and ``freezes'' high-value negative constraints, effectively pruning the search space and forcing the LLM to explore novel solutions rather than cycling through invalid ones.

\begin{table}[!t]
\centering
\small
\caption{Evolution of Cumulative Recurrent Errors (CRE) across iterations. Values represent the average count of repeated mistakes per iteration (averaged over 5 seeds). The declining trend demonstrates the system's increasing ability to avoid past failures.}
\label{tab:cre_results}
\begin{tabular}{l|ccccc}
\toprule\textbf{Task} & \textbf{Iter 1} & \textbf{Iter 2} & \textbf{Iter 3} & \textbf{Iter 4} & \textbf{Iter 5} \\\midrule
\emph{User.} & 2.20 & 1.60 & 1.00 & 0.67 & 0.33 \\
\emph{Mask.} & 1.80 & 1.67 & 1.33 & 1.00 & 0.67 \\
\emph{Mob. N$\rightarrow$N} & 2.20 & 1.40 & 1.33 & 0.67 & 0.33 \\
\emph{Mob. A$\rightarrow$A} & 2.20 & 1.60 & 1.67 & 1.00 & 0.67 \\
\emph{Mob. N$\rightarrow$A} & 2.60 & 2.00 & 2.00 & 1.33 & 0.67 \\
\midrule
\rowcolor{blue!15} \textbf{Average} & \textbf{2.20} & \textbf{1.65} & \textbf{1.47} & \textbf{0.93} & \textbf{0.53} \\\bottomrule
\end{tabular}
\end{table}

\subsection{Dynamics of Feedback Resolution}
\label{ssec:resolution_dynamics}

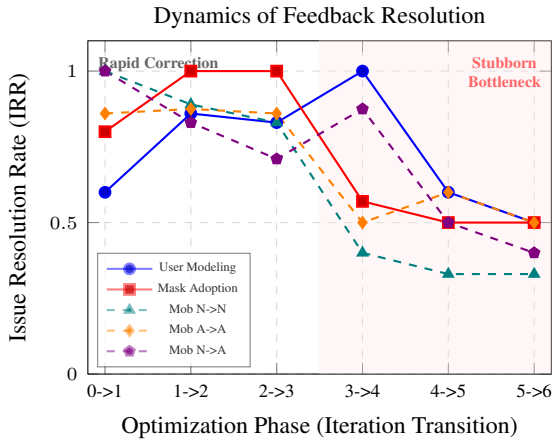
\begin{figure}[t]
    \centering
    \begin{tikzpicture}
        \begin{axis}[
            width=0.48\textwidth, 
            height=6cm,
            xlabel={Optimization Phase (Iteration Transition)},
            ylabel={Issue Resolution Rate (IRR)},
            ymin=0, ymax=1.1,
            xmin=0.8, xmax=6.2,
            xtick={1,2,3,4,5,6},
            xticklabels={0->1, 1->2, 2->3, 3->4, 4->5, 5->6},
            grid=major,
            grid style={dashed, gray!30},
            legend style={
                at={(0.02,0.02)}, 
                anchor=south west,
                font=\tiny,
                nodes={scale=0.8, transform shape},
                fill=white, fill opacity=0.8, draw opacity=0.5
            },
            tick label style={font=\scriptsize},
            label style={font=\small},
            title={Dynamics of Feedback Resolution},
            title style={font=\small, yshift=-1ex}
        ]

        \draw [draw=none, fill=red!10, opacity=0.3] (axis cs:3.5,0) rectangle (axis cs:6.5,1.1);
        \node[anchor=north east, font=\tiny, align=center, color=red!60] at (axis cs:6.2, 1.08) {\textbf{Stubborn}\\\textbf{Bottleneck}};
        
        \node[anchor=north west, font=\tiny, align=center, color=black!60] at (axis cs:0.8, 1.08) {\textbf{Rapid Correction}};

        \addplot+[thick, mark=*, blue, solid] coordinates {
            (1, 0.60) (2, 0.86) (3, 0.83) (4, 1.00) (5, 0.60) (6, 0.50)
        }; \addlegendentry{User Modeling}

        \addplot+[thick, mark=square*, red, solid] coordinates {
            (1, 0.80) (2, 1.00) (3, 1.00) (4, 0.57) (5, 0.50) (6, 0.50)
        }; \addlegendentry{Mask Adoption}

        \addplot+[thick, mark=triangle*, teal, dashed, mark options={solid}] coordinates {
            (1, 1.00) (2, 0.89) (3, 0.83) (4, 0.40) (5, 0.33) (6, 0.33)
        }; \addlegendentry{Mob N->N}

        \addplot+[thick, mark=diamond*, orange, dashed, mark options={solid}] coordinates {
            (1, 0.86) (2, 0.875) (3, 0.86) (4, 0.50) (5, 0.60) (6, 0.50)
        }; \addlegendentry{Mob A->A}

        \addplot+[thick, mark=pentagon*, violet, dashed, mark options={solid}] coordinates {
            (1, 1.00) (2, 0.83) (3, 0.71) (4, 0.875) (5, 0.50) (6, 0.40)
        }; \addlegendentry{Mob N->A}

        \end{axis}
    \end{tikzpicture}
    \caption{Visualizing the Dynamics of Issue Resolution Rate (IRR). A distinct phase shift is observed after Iteration 3, where the resolution rate drops as the system confronts complex structural bottlenecks.}
    \label{fig:irr_dynamics}
\end{figure}

To further investigate the efficiency of the optimization loop, we analyze the \textbf{Issue Resolution Rate (IRR)}, defined as the proportion of diagnostic issues identified in iteration $t$ that are successfully resolved in iteration $t+1$. 
This metric serves as a proxy for the ``fixability'' of errors and the efficacy of the agent's reasoning during code refinement.

\paragraph{Two-Phase Optimization Process.}
As illustrated in Figure~\ref{fig:irr_dynamics}, the optimization process exhibits two distinct phases:
\begin{enumerate}
    \item \textbf{Rapid Correction Phase (Iter $0 \to 3$):} The system demonstrates high resolution efficacy, with IRR values consistently exceeding \textbf{80\%} (e.g., reaching 100\% in User Modeling and Mask Adoption). In this phase, the feedback primarily targets ``low-hanging fruits''—syntax errors, API misuses, and obvious logical inconsistencies. The high IRR confirms that \ofsocia \emph{Code Gen Agent} can accurately interpret and act upon explicit diagnostic feedback.
    \item \textbf{Stubborn Bottleneck Phase (Iter $3 \to 6$):} As the simulation fidelity improves, the IRR drops significantly (averaging below \textbf{50\%} in Mobility tasks). The remaining issues in this phase are typically \emph{stubborn structural conflicts} (e.g., satisfying both JSD and WD metrics simultaneously in Mobility N$\rightarrow$N) or complex causal mechanisms. The lower resolution rate indicates that the agent is engaging in a difficult search for global optima within a highly constrained solution space, where fixing one issue may inadvertently trigger another (trade-offs), rather than simply failing to understand the instruction.
\end{enumerate}

\paragraph{Task Complexity Correlation.}
The IRR trajectory also reflects task difficulty. 
\textbf{User Modeling}, being a relatively simpler regression task, maintains a high resolution rate (e.g., 7/7 in Iter $3\to4$) throughout the process. 
In contrast, the \textbf{Mobility (N$\rightarrow$N)} task sees a sharp decline in resolution efficiency (dropping to 2/5 in Iter $3\to4$), highlighting the intrinsic difficulty of modeling high-dimensional spatiotemporal trajectories where latent factors (Pattern, Persona) are entangled.

\paragraph{Relationship with Recurrent Errors.}
It is crucial to distinguish the low late-stage IRR from the low Cumulative Recurrent Errors (CRE) reported in Appendix~\ref{app:cre_analysis}. 
A low CRE implies the agent \emph{does not repeat previously failed strategies}; combined with a low IRR, this suggests the agent is actively exploring \emph{novel but unsuccessful} hypotheses to solve stubborn problems. 
This characterizes a system that is robust against regression (forgetting) but honestly constrained by the intrinsic hardness of the scientific modeling task.

\begin{table*}[t]
\centering
\small
\caption{Evaluation results on the Mask Adoption task. Values are RMSE (mean $\pm$ 95\% CI), where lower is better. The best iteration for each backbone is highlighted in \textbf{bold}.}
\label{tab:backbone-results}
\setlength{\tabcolsep}{6pt}
\renewcommand{\arraystretch}{1.12}
\resizebox{\textwidth}{!}{
\begin{tabular}{lcccccc}
\toprule
\multicolumn{7}{c}{\textbf{Mask RMSE} $\downarrow$} \\
\midrule
\textbf{Backbone} & \textbf{Iter 0} & \textbf{Iter 1} & \textbf{Iter 2} & \textbf{Iter 3} & \textbf{Iter 4} & \textbf{Iter 5} \\
\midrule
Llama-3.3-70B  & 0.746$\pm$0.016 & 0.743$\pm$0.012 & \textbf{0.739$\pm$0.014} & 0.742$\pm$0.017 & 0.743$\pm$0.019 & 0.742$\pm$0.014 \\
Qwen3-Next-80B & 0.322$\pm$0.007 & \textbf{0.049$\pm$0.008} & 0.073$\pm$0.004 & 0.311$\pm$0.009 & 0.069$\pm$0.008 & 0.093$\pm$0.006 \\
GPT-5.1        & 0.134$\pm$0.010 & 0.109$\pm$0.005 & 0.097$\pm$0.011 & \textbf{0.073$\pm$0.010} & 0.080$\pm$0.009 & 0.075$\pm$0.011 \\
\bottomrule
\end{tabular}
}
\end{table*}

\begin{table}[t]
\centering
\small
\caption{Diagnostic metrics across backbone models on the Mask Adoption task. CRE denotes Cumulative Recurrent Errors (lower is better), and IRR denotes Issue Resolution Rate (higher is better). Best values in each column are highlighted in \textbf{bold}.}
\label{tab:backbone-diagnostics}
\setlength{\tabcolsep}{4pt}
\renewcommand{\arraystretch}{1.12}
\resizebox{\columnwidth}{!}{
\begin{tabular}{llccccc}
\toprule
\textbf{Metric} & \textbf{Backbone} & \textbf{0$\rightarrow$1} & \textbf{1$\rightarrow$2} & \textbf{2$\rightarrow$3} & \textbf{3$\rightarrow$4} & \textbf{4$\rightarrow$5} \\
\midrule
\multirow{3}{*}{\textbf{CRE} $\downarrow$}
& Llama-3.3-70B   & 1.80 & 1.40 & 1.00 & \textbf{0.60} & \textbf{0.40} \\
& Qwen3-Next-80B  & \textbf{1.20} & \textbf{1.00} & \textbf{0.80} & \textbf{0.60} & \textbf{0.40} \\
& GPT-5.1         & 1.80 & 1.67 & 1.33 & 1.00 & 0.67 \\
\midrule
\multirow{3}{*}{\textbf{IRR} $\uparrow$}
& Llama-3.3-70B   & 0.500 & 0.247 & 0.125 & 0.117 & 0.183 \\
& Qwen3-Next-80B  & \textbf{0.910} & 0.677 & 0.910 & \textbf{0.793} & \textbf{0.713} \\
& GPT-5.1         & 0.800 & \textbf{1.000} & \textbf{1.000} & 0.567 & 0.500 \\
\bottomrule
\end{tabular}
}
\end{table}

\subsection{Open-Source Backbone Model Evaluation}
\label{ssec:open-backbone}
This subsection provides the detailed evidence for the backbone portability analysis in \S~\ref{sec:backbone-portability}.
We replace GPT-5.1 in the same iterative code-evolution pipeline with two open-source instruction-tuned LLMs, \textit{Llama-3.3-70B-Instruct-Turbo} and \textit{Qwen3-Next-80B-A3B-Instruct}, while keeping the task setting, Blueprint construction, playbook mechanism, verification protocol, and iteration budget unchanged. We conduct this study on the \textbf{Mask Adoption} task, and report task-level RMSE over the full evolution trajectory, framework-level diagnostics for recurrent-error suppression and issue resolution, and a qualitative comparison of the playbook entries generated by different backbones.

In the remainder of this section, we use \textit{Llama-3.3-70B} and \textit{Qwen3-Next-80B} as shorthand for \textit{Llama-3.3-70B-Instruct-Turbo} and \textit{Qwen3-Next-80B-A3B-Instruct}, respectively.

We report two complementary views of performance. First, Table~\ref{tab:backbone-results} presents task-level RMSE over the full evolution trajectory (Iter~0 to Iter~5). Second, Table~\ref{tab:backbone-diagnostics} reports two framework-level diagnostics: \textit{Cumulative Recurrent Errors} (CRE; lower is better), which measures how effectively the loop suppresses repeated mistakes, and \textit{Issue Resolution Rate} (IRR; higher is better), which measures how often detected issues are successfully resolved in the subsequent refinement step.

At the task level, the results show that \socia is applicable beyond a single proprietary backbone. In particular, \textit{Qwen3-Next-80B} reaches the best RMSE of \textbf{0.049 $\pm$ 0.008} at Iter~1, outperforming the best GPT-5.1 result of \textbf{0.073 $\pm$ 0.010} at Iter~3. Qwen also maintains strong performance in later iterations (e.g., 0.069 $\pm$ 0.008 at Iter~4), indicating that the framework can deliver competitive simulator quality with an open model. This suggests that \ofsocia gains are not exclusive to a proprietary backbone.

At the framework level, both open-source models exhibit a monotonic decrease in CRE across iterations. Llama-3.3-70B decreases from 1.8 to 0.4, and Qwen3-Next-80B decreases from 1.2 to 0.4. This pattern is consistent with the main experiments and indicates that the combination of Blueprint anchoring, playbook memory, and execution-based verification consistently reduces the recurrence of previously observed errors, even when the backbone is changed. In this sense, the non-repetition pressure induced by \socia generalizes across backbone choices.

At the same time, the results also reveal that backbone capability affects how well the framework converts diagnosis into effective refinement.
Although Llama-3.3-70B also shows decreasing CRE, its RMSE remains poor throughout the trajectory (around 0.74), and its IRR is consistently the lowest among the three backbones.
This suggests that the framework still discourages repeated errors under Llama, but the model is substantially less effective at converting diagnosed issues into high-quality and executable repairs.
By contrast, Qwen3-Next-80B achieves the strongest IRR in most transitions, which aligns with its superior RMSE and suggests that higher-quality issue diagnosis and more actionable playbook entries lead to more effective refinement.

\paragraph{Qualitative analysis of playbook strategies.}
To better understand the performance gap across open-source backbones, we further inspect representative playbook entries generated during the refinement process. The qualitative patterns align closely with the diagnostic results in Table~\ref{tab:backbone-diagnostics}. 
As shown in Table~\ref{tab:playbook-examples}, Llama-3.3-70B tends to produce coarse and underspecified strategies that identify the surface symptom but provide limited mechanistic guidance for code revision. 
By contrast, Qwen3-Next-80B produces substantially higher-resolution entries that explicitly connect observed failures to concrete implementation errors and prescribe targeted refactors. 
This difference helps explain why both models exhibit decreasing CRE, yet only Qwen consistently converts identified issues into effective repairs, as reflected by its much stronger IRR.

\begin{table*}[t]
\centering
\small
\caption{Representative playbook entries generated by different open-source backbones on the Mask Adoption task. Llama-3.3-70B tends to produce coarse and underspecified strategies, whereas Qwen3-Next-80B generates more mechanistic and implementation-oriented guidance.}
\label{tab:playbook-examples}
\renewcommand{\arraystretch}{1.15}
\setlength{\tabcolsep}{5pt}
\begin{tabularx}{\textwidth}{>{\raggedright\arraybackslash}p{2.6cm} >{\raggedright\arraybackslash}X >{\raggedright\arraybackslash}X}
\toprule
\textbf{Field} & \textbf{Llama-3.3-70B} & \textbf{Qwen3-Next-80B} \\
\midrule
\textbf{Strategy name} 
& \texttt{missing-neighbor-mask-share-calculation}
& \texttt{zero-layer-weights-calibrated} \\
\midrule
\textbf{Error identification}
& The simulator is underperforming due to inaccurate neighbor mask share calculation.
& All peer influence and layer weight parameters (\texttt{w\_*}, \texttt{beta\_*}, \texttt{phi\_*}) are calibrated to zero, causing the model to ignore social network structure and rely solely on risk perception and broadcast, resulting in poor aggregate and transition fit. \\
\midrule
\textbf{Root cause analysis}
& The neighbor mask share calculation in the \texttt{Simulator.rollout} method is incorrect.
& In \texttt{GradientBasedCalibrator.fit}, the feature vector incorrectly includes both layer-weighted shares (\texttt{w\_f * share\_f}) and peer-slope terms (\texttt{beta\_f * share\_f}) as separate additive terms, but the optimization penalizes non-zero weights via L2 regularization and finds the minimal-loss solution by driving all peer-related coefficients to zero due to collinearity and lack of constraint on layer-weight normalization during optimization. \\
\midrule
\textbf{Correct approach}
& Update the neighbor mask share calculation to correctly account for the previous day's state.
& Remove the \texttt{w\_family}, \texttt{w\_work}, and \texttt{w\_community} parameters from the feature vector and decision rule; use only \texttt{beta\_f}, \texttt{beta\_w}, and \texttt{beta\_c} as direct peer influence coefficients. Enforce layer-weight normalization (sum-to-one) only for interpretation, not as learnable parameters. Update \texttt{Simulator.\_compute\_adoption\_probability} and \texttt{GradientBasedCalibrator.objective} to use only \texttt{beta\_*} terms for peer influence. \\
\bottomrule
\end{tabularx}
\end{table*}

\paragraph{Summary.}
Overall, these results show that \socia generalizes beyond a single proprietary backbone. Open-source models can preserve the same framework-level error-reduction dynamics, and sufficiently capable ones can also achieve highly competitive task performance. 
The remaining gap is explained by differences in playbook quality and issue-to-repair execution fidelity.

\section{Qualitative Analysis: Structural Evolution on Mob. N$\to$A}
\label{append:qualitative_n2a}

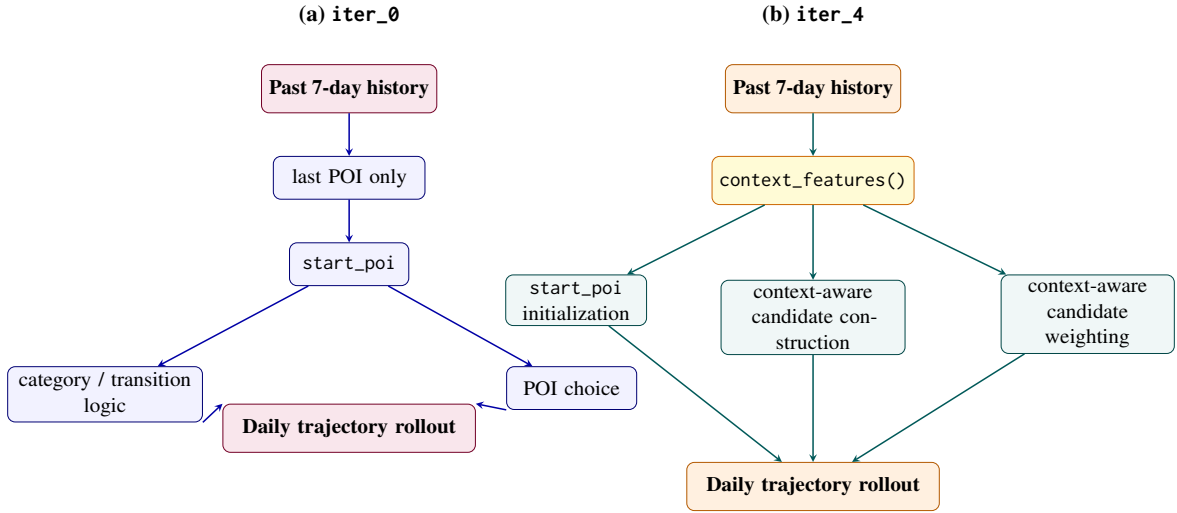
\begin{figure*}[t]
\centering
\resizebox{0.96\textwidth}{!}{%
\begin{tikzpicture}[
    node distance=8mm and 10mm,
    title/.style={font=\bfseries\large},
    box0/.style={
        draw=blue!50!black,
        fill=blue!8,
        rounded corners,
        align=center,
        minimum height=9mm,
        inner sep=4pt
    },
    smallbox0/.style={
        draw=blue!45!black,
        fill=blue!5,
        rounded corners,
        align=center,
        minimum height=8mm,
        inner sep=3pt,
        text width=26mm
    },
    box4/.style={
        draw=teal!60!black,
        fill=teal!10,
        rounded corners,
        align=center,
        minimum height=9mm,
        inner sep=4pt
    },
    smallbox4/.style={
        draw=teal!55!black,
        fill=teal!6,
        rounded corners,
        align=center,
        minimum height=8mm,
        inner sep=3pt,
        text width=32mm
    },
    emph0/.style={
        draw=purple!60!black,
        fill=purple!10,
        rounded corners,
        align=center,
        minimum height=9mm,
        inner sep=4pt
    },
    emph4/.style={
        draw=orange!70!black,
        fill=orange!12,
        rounded corners,
        align=center,
        minimum height=9mm,
        inner sep=4pt
    },
    arrow0/.style={->, >=stealth, thick, draw=blue!65!black},
    arrow4/.style={->, >=stealth, thick, draw=teal!70!black}
]

\node[title] (title0) at (0,0) {(a) \texttt{iter\_0}};

\node[emph0, below=6mm of title0] (hist0) {\textbf{Past 7-day history}};
\node[smallbox0, below=8mm of hist0] (last0) {last POI only};
\node[smallbox0, below=8mm of last0, text width=20mm] (start0) {\texttt{start\_poi}};

\node[smallbox0, below left=15mm and 16mm of start0, text width=34mm] (trans0) {category / transition\\logic};
\node[smallbox0, below right=15mm and 18mm of start0, text width=22mm] (poi0) {POI choice};

\node[emph0, below=22mm of start0, text width=44mm] (day0) {\textbf{Daily trajectory rollout}};

\draw[arrow0] (hist0) -- (last0);
\draw[arrow0] (last0) -- (start0);
\draw[arrow0] (start0) -- (trans0);
\draw[arrow0] (start0) -- (poi0);
\draw[arrow0] (trans0.south east) -- (day0.north west);
\draw[arrow0] (poi0.south west) -- (day0.north east);

\node[title] (title4) at (8.6,0) {(b) \texttt{iter\_4}};

\node[emph4, below=6mm of title4] (hist4) {\textbf{Past 7-day history}};
\node[
    draw=orange!80!black,
    fill=yellow!20,
    rounded corners,
    align=center,
    minimum height=9mm,
    inner sep=4pt
] (ctx4) [below=8mm of hist4] {\texttt{context\_features()}};

\node[smallbox4, below left=13mm and 12mm of ctx4, text width=24mm] (start4) {\texttt{start\_poi}\\initialization};
\node[smallbox4, below=14mm of ctx4, text width=32mm] (cand4) {context-aware\\candidate construction};
\node[smallbox4, below right=13mm and 16mm of ctx4, text width=30mm] (weight4) {context-aware\\candidate weighting};

\node[emph4, below=20mm of cand4, text width=44mm] (day4) {\textbf{Daily trajectory rollout}};

\draw[arrow4] (hist4) -- (ctx4);
\draw[arrow4] (ctx4) -- (start4);
\draw[arrow4] (ctx4) -- (cand4);
\draw[arrow4] (ctx4) -- (weight4);
\draw[arrow4] (start4) -- (day4);
\draw[arrow4] (cand4) -- (day4);
\draw[arrow4] (weight4) -- (day4);

\end{tikzpicture}%
}
\caption{Interaction-topology schematic for the mobility simulator. In \texttt{iter\_0}, recent history only seeds the first POI of the day. In \texttt{iter\_4}, the same 7-day history is transformed into reusable context features that influence start-of-day initialization, candidate construction, and candidate weighting throughout the full daily rollout. This illustrates a mechanism-level structural refinement induced by iterative feedback.}
\label{fig:mobility-context-topology}
\end{figure*}

We qualitatively analyze why the Mob. N$\to$A simulator improves from \texttt{iter\_0} to \texttt{iter\_4}. 
The key driver is \emph{structural alignment} between the simulator and the evaluator with the Blueprint, rather than merely increasing calibration iterations.
Across iterations, POI hit-based metrics (e.g., recall) steadily improve, transition-level divergence consistently decreases toward its best values, and spatial-distance discrepancies shrink substantially, while the overall objective reaches its best performance in the final iteration.
Below we trace this trajectory through the evolution of code structure and modeling choices.

\subsection{Iteration-by-Iteration Code Evolution}
\label{ssec:n2a_iter_evolution}

\paragraph{\texttt{iter\_0}: End-to-end runnable, but partially unoptimizable.}
The initial implementation establishes the full pipeline---data split, calibration, rollout, and metric computation---but several core components remain fragile.
In particular, trajectory parsing/serialization and token matching are insufficiently canonicalized, causing downstream metrics to degrade into uninformative regimes (e.g., recall collapses, transition metrics saturate, and distance metrics become unreliable due to failed coordinate joins).
As a result, the calibrator is effectively optimizing an objective polluted by representation artifacts rather than genuine behavioral discrepancies.

\paragraph{\texttt{iter\_1}: Make the evaluator trustworthy before improving the simulator.}
The first major step is to repair the \emph{measurement layer}: trajectory parsing, strict string formatting, and metric implementations (notably recall and transition metrics) are fixed so that the objective reflects real differences between simulated and observed behavior.
This shift is crucial: once the evaluator becomes consistent with the dataset schema, calibration becomes directionally meaningful.
A notable side effect is that some errors appear larger after this repair---not because the simulator worsened, but because the evaluation stopped masking mismatches and began exposing true deficits that were previously hidden by broken scoring.

\paragraph{\texttt{iter\_2}: Enforce representation invariances and stabilize temporal modeling.}
With evaluators corrected, the next structural upgrade is robust \emph{canonicalization}:
POI tokens are normalized (e.g., punctuation/whitespace handling) to reduce spurious mismatches, and time-of-day modeling is made consistent by adopting an explicit binning strategy that is shared by both generation and evaluation.
These changes reduce variance in the objective and improve comparability across days and users, making the calibration landscape smoother.
In practice, this iteration marks the transition from ``fixing bugs'' to ``reducing estimator noise''.

\paragraph{\texttt{iter\_3}: Strengthen behavioral conditionality and move from mean-matching to distribution fitting.}
The third iteration upgrades the \emph{generative mechanism} itself.
POI choice becomes strongly conditional on super-category (or an equivalent high-level activity type), while incorporating per-user preference signals and anchor reuse to enhance personalization.
In parallel, stop-count modeling is improved from coarse heuristics (often mean-driven) to distribution-aware fitting, so the simulator can match not only expected counts but also the shape of the stop-count distribution.
These structural changes are reflected in substantial improvements in hit-based measures and transition consistency, indicating that the simulator begins to reproduce user-specific mobility signatures rather than generic category-level patterns.

\paragraph{\texttt{iter\_4}: Inject 7-day context throughout the day, enforce time-budget realism, and correct spatial bias.}
The final iteration integrates multiple missing ingredients into a coherent design and yields the best overall behavior.
First, the 7-day context is elevated from a weak initialization cue to a \emph{persistent conditioning signal} that influences decisions throughout the day by injecting recent POIs into candidate sets, improving both personalization and POI-level fidelity.
Second, the simulator introduces explicit \emph{time-budget} constraints for stop and gap sampling, preventing unrealistic truncation artifacts and stabilizing time-related statistics.
Third, spatial evaluation is revised to explicitly account for coordinate missingness, computing distance-based metrics on resolvable step pairs to avoid biased comparisons.
Finally, previously restrictive parameter bounds (e.g., clipped mixture weights between preference and distance) are relaxed to permit calibration to explore the full admissible range, improving parameter identifiability.
Together, these changes produce the strongest improvements in POI hit-based metrics, the lowest transition divergences, the best spatial-distance alignment, and the best overall objective.

\subsection{Code-Level Structural Mechanism Change: From Start-Only Context to Persistent Daily Conditioning}
\label{ssec:mobility-structural-change}

We compare the simulator structures of the earlier version (\texttt{iter\_0}) and the refined version (\texttt{iter\_4}) through both code-level differences (Fig.~\ref{fig:mobility-before-after-code}) and interaction-topology changes (Fig.~\ref{fig:mobility-context-topology}) to analyze how our method iteratively improves simulator structure through feedback from numerical analysis.

This code difference also induces a corresponding change in the simulator's interaction topology. As illustrated in Fig.~\ref{fig:mobility-context-topology}, the earlier simulator uses recent history only at the entry point of the rollout, namely to seed the first POI, after which the subsequent trajectory is governed by transition/category logic alone. In contrast, the refined simulator routes the same 7-day history through reusable context features that influence not only start-of-day initialization, but also candidate construction and candidate weighting throughout the full daily rollout. In this sense, the refined simulator changes the effective information-flow topology between past behavior and within-day decision making, rather than merely adjusting coefficients within an otherwise fixed mechanism.

This refinement is also consistent with the broader qualitative trajectory reported in our appendix analysis: \texttt{iter\_4} improves by elevating the 7-day context from a weak initialization cue to a persistent conditioning mechanism that influences decisions throughout the day, thereby improving personalization and POI-level fidelity. Taken together, Fig.~\ref{fig:mobility-before-after-code} and Fig.~\ref{fig:mobility-context-topology} show that the improvement is not merely that the refined simulator uses different calibrated parameters, but that it reorganizes the simulator's information flow. Numerical feedback identifies persistent fidelity errors, and the outer-loop structural revision responds by relocating recent-history signals from a single initialization point to a mechanism that conditions decisions across the entire day.

\subsection{Takeaway: A Three-Stage Maturation Pattern}
\label{ssec:n2a_takeaway}

The Mob. N$\to$A evolution exhibits a clear three-stage maturation pattern:
(i) \emph{Make optimization well-posed} by fixing parsing, serialization, and evaluation mismatches that would otherwise cause calibration to chase spurious artifacts;
(ii) \emph{Make modeling internally consistent} through token canonicalization, blueprint-aligned metrics, and shared temporal abstractions between simulation and evaluation;
(iii) \emph{Make behavior mechanism-faithful} by introducing persistent 7-day context conditioning, explicit time-budget realism, bias-aware spatial scoring, and fully calibratable behavioral degrees of freedom.
Taken together, these refinements transform calibration from optimizing noisy implementation artifacts into optimizing distributional fidelity, which explains why later iterations progressively converge toward higher-fidelity simulator behavior.

\begin{figure}[!htbp]
\centering
\setlength{\abovecaptionskip}{4pt}
\setlength{\belowcaptionskip}{-2pt}

\begin{minipage}[t]{\columnwidth}
\centering
\textbf{Before (\texttt{iter\_0}): start-only context}
\vspace{2pt}
\begin{lstlisting}[language=Python,basicstyle=\ttfamily\scriptsize,columns=fullflexible,breaklines=true]
def simulate_day(..., context_last_poi, ...):
    if context_last_poi is not None:
        start_poi = context_last_poi
    elif self.baseline.anchor_pois:
        start_poi = weighted_choice(
            self.baseline.anchor_pois,
            self.baseline.anchor_weights,
            rng.u()
        )
    else:
        start_poi = sim.global_popular_pois[0]

    ...
    for sc in sc_seq:
        candidates = sim.dataset.poi_tokens_by_supercat.get(sc)
        if not candidates:
            candidates = sim.global_popular_pois[:5000]
        ...
\end{lstlisting}
\end{minipage}

\vspace{0.2em}

\begin{minipage}[t]{\columnwidth}
\centering
\textbf{After (\texttt{iter\_4}): persistent daily conditioning}
\vspace{2pt}
\begin{lstlisting}[language=Python,basicstyle=\ttfamily\scriptsize,columns=fullflexible,breaklines=true]
def simulate_day(..., context_last_poi, ...):
    ctx_tok = sim.resolve_poi_token(context_last_poi) \
        if context_last_poi is not None else None
    ctx = sim.context_features(self.user_id, date_obj,
                               context_days=7)
    ctx_top_any = ctx.get("top_pois_any", [])
    ctx_counts_by_sc = ctx.get("counts_by_supercat", {})
    ctx_probs_by_sc = ctx.get("probs_by_supercat", {})

    if ctx_tok is not None:
        start_poi = ctx_tok
    elif ctx_top_any:
        start_poi = str(ctx_top_any[0])
    elif self.baseline.anchor_pois:
        start_poi = weighted_choice(...)
    else:
        start_poi = sim.global_popular_pois[0]

    ...
    for sc in sc_seq:
        ctx_top_sc = ...
        user_top = ...
        global_top = ...
        catalog_sample = ...
        candidates = list(dict.fromkeys(
            ctx_top_sc + user_top + global_top + catalog_sample
        ))
        ...
\end{lstlisting}
\end{minipage}

\caption{Code-level mechanism refinement in the mobility simulator. \texttt{iter\_0} uses recent history only to seed the first POI, while \texttt{iter\_4} converts the same 7-day history into reusable context features that condition candidate construction throughout the daily rollout.}
\label{fig:mobility-before-after-code}
\end{figure}

\FloatBarrier

\section{Strategy Playbook: Structure and Contents}
\label{sec:playbook_structure}

In this section, we present an example Strategy Playbook entry for a remedial strategy and analyze its structure as well as the role of each field.

\socia maintains a \emph{Strategy Playbook} as a persistent repository of remediation knowledge accumulated across iterations.
The Playbook is a JSON artifact with two top-level components: \texttt{playbook\_metadata} and \texttt{strategies}.
We describe each component below.

\subsection{Playbook Metadata}
\label{subsec:playbook_metadata}

The \texttt{playbook\_metadata} block summarizes the global state of the Playbook at the time it is saved.
It includes:
(i) versioning and provenance (\texttt{version}, \texttt{project\_name}); 
(ii) recency information (\texttt{last\_updated\_time}, \texttt{last\_updated\_iteration});
(iii) size statistics (\texttt{total\_token\_count}, \texttt{total\_insights}); and
(iv) outcome bookkeeping (\texttt{solved\_count}, \texttt{unsolved\_count}, \texttt{deleted\_count}, \texttt{finalized\_at}).
This metadata provides a compact snapshot for memory budgeting (e.g., token counts) and for monitoring the health of the knowledge base (e.g., unresolved items).

\subsection{Strategy Entries}
\label{subsec:strategy_entries}

The core content lives in \texttt{strategies}, a dictionary keyed by a unique \emph{strategy id} (a human-readable slug).
Each strategy entry comprises two parts:
\texttt{meta\_info} (usage- and status-level bookkeeping) and \texttt{reflection} (the actionable, evidence-grounded remediation record).

\paragraph{Meta-information (\texttt{meta\_info}).}
\texttt{meta\_info} captures how a strategy behaves over time:
\texttt{token\_count} measures its prompt footprint;
\texttt{status} indicates whether it is currently considered \texttt{resolved} or not;
\texttt{usage\_count} and \texttt{unusage\_count} record retrieval/selection frequency;
\texttt{success\_attribution} and \texttt{failure\_attribution} track how often the strategy is empirically associated with improvements or regressions.
These fields support reliability-aware selection under a fixed context budget.

\paragraph{Reflection record (\texttt{reflection}).}
\texttt{reflection} is the substantive content of a strategy: it formalizes a repair as a testable hypothesis grounded in metrics and traceability links.
It contains:
\begin{itemize}
    \item \textbf{Problem typing and priority:} \texttt{issue\_type}, \texttt{severity}, and \texttt{from\_user\_feedback}.
    \item \textbf{Anchors to authoritative constraints:} \texttt{blueprint\_refs}, which point to the relevant requirements/definitions in the Blueprint.
    \item \textbf{Traceability to code:} \texttt{code\_refs}, which list implicated program symbols (and optionally line spans if available).
    \item \textbf{Evidence:} \texttt{evidence} stores supporting signals such as \texttt{metrics}, \texttt{error\_logs}, and (if present) \texttt{user\_feedback}.
    \item \textbf{Causal diagnosis:} \texttt{error\_identification} states what is wrong; \texttt{root\_cause\_analysis} explains why it happens.
    \item \textbf{Actionable remediation:} \texttt{correct\_approach} specifies concrete corrective steps to apply in subsequent iterations.
    \item \textbf{Portable lesson:} \texttt{key\_insight} distills a generalizable principle beyond a single patch.
    \item \textbf{Metric grounding:} \texttt{metric\_links} enumerates which metrics the strategy targets, including directionality and weights.
\end{itemize}

\subsection{Example Strategy (Illustrative Field Semantics)}
\label{subsec:playbook_example}

To illustrate how a single entry is structured, consider a resolved strategy whose id indicates a mismatch in stop-count modeling.
Its reflection:
(i) classifies the issue as an evaluation signal with high severity;
(ii) links to Blueprint items that define the stop-count metric and the associated calibratable parameter;
(iii) references the implicated code symbols in the simulator and evaluator;
(iv) records metric evidence showing systematic stop-count deviations and distribution mismatch;
(v) diagnoses the root cause as a mismatch between sampled sequence length and later truncation during time-feasibility checks; and
(vi) proposes a corrective approach that (a) applies a distribution-aware length adjustment (rather than rounding counts) and (b) enforces time feasibility during gap sampling to prevent truncation-induced bias.
Finally, the entry ties the remediation to explicit target metrics via \texttt{metric\_links}, enabling the system to prioritize strategies that directly address the current failure modes.

\paragraph{Playbook Fields.}
\begin{itemize}
  \item \texttt{playbook\_metadata.*}: Global snapshot of the Playbook, including version, recency, size, and solved/unsolved bookkeeping.
  \item \texttt{strategies[\emph{id}].meta\_info}: Retrieval footprint and reliability signals (e.g., usage counts, attribution, and status).
  \item \texttt{issue\_type}/\texttt{severity}: The failure category and its urgency level.
  \item \texttt{blueprint\_refs}: Which Blueprint constraints/definitions this strategy must respect.
  \item \texttt{code\_refs}: The implicated code symbols for debugging and patching.
  \item \texttt{evidence}: Empirical signals supporting the hypothesis (metrics, logs, and/or user feedback).
  \item \texttt{error\_identification}: The observed discrepancy (what goes wrong).
  \item \texttt{root\_cause\_analysis}: Why the discrepancy occurs in the implementation/design.
  \item \texttt{correct\_approach}: Concrete repair steps to apply in the next iteration.
  \item \texttt{key\_insight}: A transferable lesson that generalizes across iterations/tasks.
  \item \texttt{metric\_links}: The target metrics and their relative weights/directions.
\end{itemize}

\begin{tcolorbox}[
  enhanced,
  breakable,
  width=\linewidth,
  colback=gray!10,
  colframe=red!70!black,
  title={\textbf{Illustration of the playbook.}},
  fonttitle=\bfseries,
  boxrule=0.8pt,
  arc=0pt,
  left=3pt,
  right=3pt,
  top=3pt,
  bottom=3pt,
  boxsep=2pt,
  before upper={\setlength{\parindent}{0pt}\ttfamily\footnotesize}
]

{\scriptsize
\begin{lstlisting}[breaklines=true, breakatwhitespace=true, columns=fullflexible, frame=none]
{
  {
  "playbook_metadata": {
    "version": "v0.1",
    "project_name": "",
    "last_updated_time": "2026-01-04T10:18:58.021231",
    "last_updated_iteration": "3",
    "total_token_count": 1555,
    "total_insights": 34,
    "solved_count": 28,
    "unsolved_count": 6,
    "deleted_count": 0,
    "finalized_at": "2026-01-03T18:07:02.323788"
  },
  "strategies": {
    "stop-count-model-underestimates-
    and-misaligns-distribution": {
      "meta_info": {
        "token_count": 339,
        "status": "resolved",
        "usage_count": 1,
        "unusage_count": 0,
        "success_attribution": 1,
        "failure_attribution": 0
      },
      "reflection": {
        "issue_type": "EVAL_SIGNAL",
        "severity": "high",
        "from_user_feedback": false,
        "blueprint_refs": [
          "stop_count_mae definition",
          "theta_stop_count_multiplier calibratable parameter"
        ],
        "code_refs": [
          {
            "symbol": "Resident.simulate_day",
            "lines": "unknown"
          },
          {
            "symbol": "Evaluator.compute_metrics",
            "lines": "unknown"
          }
        ],
        "evidence": {
          "user_feedback": null,
          "error_logs": null,
          "metrics": "stop_count_abs_mean_error=2.22; diagnostics gt_mean_visits_per_day=7.759 vs sim_mean_visits_per_day=6.686; stop_count_kl=0.4452"
        },
        "error_identification": "Stop counts are materially off: simulations average ~1.07 fewer visits/day than ground truth, with large per-day absolute error (~2.22) and distribution mismatch (KL~0.45).",
        "root_cause_analysis": "Stop-count generation rescales a discrete pmf by rounding counts (c2 = round(c * mult)) and then applies ad-hoc neighbor smoothing (adding mass to c+/-1). This can distort the original distribution (multi-modal shapes collapse), and the later truncation in simulate_day (break if t_min >= day) further reduces realized stop count, biasing downward vs the sampled stop_count."
        "correct_approach": "Stabilize stop-count control and prevent truncation bias:\n1) Sample stop_count directly from the baseline pmf, then apply multiplier by shifting probabilities rather than rounding counts (e.g., reweight each count c by exp(-|c - c*mult|/tau) and renormalize).\n2) Track when the time loop truncates (t_min>=MINUTES_PER_DAY) and either (a) cap gap sampling to fit remaining day time, or (b) allow later-day timestamps up to 23:59 by compressing gaps.\n3) Add a diagnostic: sampled_stop_count vs realized_len(visits) to quantify truncation-induced error per split.",
        "key_insight": "If you sample a target sequence length but later time-feasibility truncates the sequence, you must either enforce feasibility during length sampling or adjust gaps to honor the sampled length; otherwise stop-count metrics will be systematically biased.",
        "metric_links": [
          {
            "name": "stop_count_abs_mean_error",
            "direction": "lower_is_better",
            "weight": 0.6
          },
          {
            "name": "stop_count_kl",
            "direction": "lower_is_better",
            "weight": 0.4
          }
        ]
      }
    }
  }
}
}
\end{lstlisting}
}
\end{tcolorbox}

\begin{center}
\footnotesize
\textit{Example of playbook and its managed strategy.}
\label{example:playbook}
\end{center}

\section{Blueprint Design}
\label{appen:blueprint}

\subsection{Blueprint Synthesis via Data-Analysis-Driven Specification}
\label{ssec:blueprint_synthesis}

\paragraph{Rationale.}
Automated simulator construction is a long-horizon search over executable programs, where early mistakes in data interpretation can propagate and dominate later iterations.
To stabilize this search, \socia introduces a \emph{Blueprint}: a structured, persistent specification that translates a task description and raw observational data into an explicit simulator design space.
The Blueprint functions as an authoritative anchor that (i) constrains what mechanisms may be introduced, (ii) specifies what is measurable and calibratable, and (iii) provides a shared reference for all downstream agents throughout iterative refinement.

\subsection{Inputs and Outputs}
\label{ssec:blueprint_io}

The Data Analysis stage takes two inputs: a natural-language task specification and a collection of observational datasets.
Its output is a machine-readable Blueprint consisting of ten tightly scoped components: (1) overall simulation design, (2) scale and granularity, (3) agent archetypes, (4) interaction topology, (5) information propagation, (6) exogenous signals, (7) action and decision policy, (8) holdout plan, (9) simulation evaluation, and (10) calibratable parameters.
Together, these components define \emph{what the simulator is}, \emph{what it can vary}, and \emph{how its fidelity will be judged}.

\subsection{From Observational Data to Simulator Design Constraints}
\label{ssec:blueprint_from_data}

\paragraph{Schema-grounded understanding.}
Rather than asking the model to invent a simulator from scratch, we first ground the analysis in \emph{what the data can support}.
For each dataset, the agent extracts lightweight structural evidence—identifiers, record formats, temporal fields, spatial keys, and linkable attributes—and distills it into a semantic summary.
These summaries establish a contract between evidence and modeling: which entities exist, what states can be observed, how time and space are represented, and which files can be joined to recover latent attributes (e.g., coordinates or categories).
By explicitly encoding these constraints, the Blueprint prevents downstream agents from hallucinating incompatible schemas or unsupported mechanisms.

\paragraph{Evidence-aligned decomposition.}
Using the task description as the objective and the data summaries as evidence, the Data Analysis agent decomposes the simulator into interpretable building blocks.
It specifies (i) the agent unit of simulation (e.g., one resident per observed user), (ii) the state representation and update dynamics consistent with observed trajectories, (iii) the interaction channels that are supported (or deliberately excluded) by the data, and (iv) exogenous signals derivable from timestamps or known covariates.
Crucially, this decomposition is \emph{not} merely descriptive: it determines what the downstream code generator is allowed to implement and what must be treated as designed assumptions versus data-derived facts.

\subsection{Calibration-Ready Parameterization}
\label{ssec:blueprint_calibration_ready}

Beyond qualitative design, the Blueprint is constructed to be \emph{calibration-ready}.
It enumerates a set of calibratable parameters (global and/or agent-specific), assigns feasible ranges or structural constraints, and links them to measurable consequences in the data (e.g., time-of-day profiles, transition tendencies, spatial jump distributions).
This explicit parameterization enables a numerical inner-loop calibrator to optimize continuous parameters while the outer loop focuses on structural and policy-level revisions.
As a result, the Blueprint enforces a principled separation between \emph{what should be tuned} and \emph{what should be rewritten}.

\subsection{Calibrator Example}
\label{ssec:calibrator_example}

As shown in the Blueprint example in \S\ref{ssec:blueprint_example}, the calibrator is not introduced in an ad-hoc manner after simulator generation, but is already specified at the task-design stage as part of the Blueprint. Concretely, the Blueprint explicitly defines a \texttt{CalibrationController} role, whose function is to propose a parameter vector $\theta$, run simulation on the 2021 validation subset, compute an objective from evaluation metrics, and update $\theta$ via a black-box optimization strategy. It also specifies the split protocol for calibration, namely that baseline fitting uses only 2019--2020 data, parameter calibration uses only the per-user 2021 \texttt{V\_calib} subset, and the held-out \texttt{V\_test} subset is reserved for final reporting. 
In addition, the Blueprint enumerates the calibratable abnormal-shift parameters together with their feasible bounds, including:
\begin{itemize}[leftmargin=1.2em, itemsep=1pt, topsep=2pt, parsep=0pt]
    \item \texttt{theta\_cat\_\allowbreak weight\_multiplier\_\allowbreak by\_\allowbreak supercategory}
    \item \texttt{theta\_stop\_count\_multiplier}
    \item \texttt{theta\_start\_time\_shift\_minutes}
    \item \texttt{theta\_distance\_decay\_scale}
    \item \texttt{theta\_preference\_vs\_distance\_mixture}
    \item \texttt{theta\_infrastructure\_stop\_bonus}
\end{itemize}
This means that the Blueprint does not merely describe the simulator qualitatively; it already specifies what should be optimized, under what constraints, and against which validation signal.

The simulator code generated in \texttt{iter\_4} then concretizes this Blueprint-level calibrator design into an executable inner-loop program.
Specifically, the generated \texttt{RandomSearchCalibrator} instantiates a bounded search space directly from the Blueprint constraints. For example, the stop-count multiplier, start-time shift, distance-decay scale, preference--distance mixture, and infrastructure-stop bonus are sampled from their corresponding feasible ranges, while each super-category multiplier is sampled independently from its Blueprint-prescribed interval. In this way, the search space is not manually guessed in code, but instantiated from the parameter schema already specified in the Blueprint.

A short excerpt of the generated calibrator is shown below.

\begin{lstlisting}[
language=Python,
caption={Executable calibrator synthesized from Blueprint parameter bounds (abridged).},
label={lst:calibrator_example},
breaklines=true,
breakatwhitespace=false,
columns=fullflexible,
keepspaces=true,
basicstyle=\ttfamily\footnotesize
]
class RandomSearchCalibrator(Calibrator):
    def _sample_params(self, rng):
        def ru(lo, hi):
            return lo + (hi - lo) * rng.u()

        params = {
            "theta_stop_count_multiplier":
                ru(0.5, 1.8),
            "theta_start_time_shift_minutes":
                int(ru(-120, 120)),
            "theta_distance_decay_scale":
                ru(0.5, 2.5),
            "theta_preference_vs_distance_mixture":
                ru(0.0, 1.0),
            "theta_infrastructure_stop_bonus":
                ru(0.0, 3.0),
        }

        cat_mult = {}
        for sc in self.tuned_supercats:
            cat_mult[sc] = ru(0.25, 4.0)

        params[
            (
                "theta_cat_weight_multiplier_"
                "by_supercategory"
            )
        ] = cat_mult
        return params
\end{lstlisting}

The second aspect is the calibration objective. In the Blueprint, calibration is defined as optimizing a weighted objective over validation metrics on 2021 trajectories, rather than fitting to a single scalar target. This design is also realized directly in code. The generated \texttt{Evaluator} computes multiple task-level metrics, including:
\begin{itemize}[leftmargin=1.5em, topsep=2pt, itemsep=1pt, parsep=0pt]
    \item \texttt{category\_share\_mae}
    \item \texttt{stop\_count\_abs\_mean\_error}
    \item \texttt{stop\_count\_kl}
    \item \texttt{tod\_jsd\_avg}
    \item \texttt{topk\_poi\_recall}
    \item \texttt{transition\_divergence}
    \item \texttt{trip\_distance\_wasserstein}
\end{itemize}
It then aggregates them into a single weighted objective, with recall converted into a minimization term via $1-\texttt{topk\_poi\_recall}$.
The calibrator then repeatedly rolls out the simulator on \texttt{V\_calib}, evaluates the resulting trajectories, records each trial in a calibration log, and retains the parameter vector with the best validation objective.

\begin{lstlisting}[language=Python, caption={Metric aggregation used as the calibration objective.}, label={lst:calibrator_objective}]
def objective(self, metrics):
    w = self.objective_weights
    obj = 0.0
    for k, weight in w.items():
        if k == "topk_poi_recall":
            obj += float(weight) * (1.0 - float(metrics[k]))
        else:
            obj += float(weight) * float(metrics[k])
    return float(obj)
\end{lstlisting}

In this concrete task instance, the synthesized calibrator uses bounded random search; more generally, the same Blueprint-to-code translation can instantiate stronger numerical optimizers such as Bayesian optimization when required by the task.

Thus, this example makes the bi-level logic concrete. The Data Analysis Agent first produces a Blueprint that already contains a calibration-ready specification: which parameters are tunable, what their valid ranges are, what validation split must be used, and which metrics should define success. The Code Generation Agent then translates this symbolic specification into an executable calibrator program \(C_t\), whose search space is bounded by Blueprint constraints and whose loss function is induced by Blueprint metrics. In this sense, the calibrator is not a post-hoc utility attached to the simulator, but an executable realization of the calibration semantics already encoded in the Blueprint.

\subsection{Evaluation and Holdout as First-Class Design}
\label{ssec:blueprint_evaluation_holdout}

A key role of the Blueprint is to make evaluation and data splitting \emph{first-class} rather than an afterthought.
The agent specifies a temporally consistent holdout plan to prevent leakage, and defines the evaluation metrics as distributional comparisons between simulated and observed behaviors.
By encoding evaluation in the Blueprint, we ensure that every iteration of simulator synthesis is accountable to quantitative fidelity targets, and that improvements can be attributed to specific design or calibration changes.

\paragraph{Summary.}
In sum, Blueprint synthesis converts task intent and observational evidence into a structured simulator design space with explicit constraints, calibratable degrees of freedom, and evaluation criteria.
This transforms simulator construction from ad hoc code generation into a grounded scientific modeling workflow, where each downstream agent operates under an explicit, evidence-aligned contract.

\subsection{Blueprint Prompt}
\label{ssec:blueprint_prompt}

In this section, we provide the prompt used by the Data Analysis agent to generate the Blueprint.

\begin{tcolorbox}[
  enhanced,
  breakable,
  width=\linewidth,
  colback=gray!10,
  colframe=red!70!black,
  title={\textbf{Data Analysis Agent Prompt}},
  fonttitle=\bfseries,
  boxrule=0.8pt,
  arc=0pt,
  left=3pt,
  right=3pt,
  top=3pt,
  bottom=3pt,
  boxsep=2pt,
  before upper={\setlength{\parindent}{0pt}\ttfamily\footnotesize}
]
You are an expert data scientist and simulation modeler. Your task is to analyze the provided task description and data summaries to design a simulation model blueprint for subsequent code generation.

Overall simulation design: automatically generate and calibrate the simulator. This requires defining the classes and functions within the simulator, as well as how these classes and functions interact.

\medskip

\textbf{TASK DESCRIPTION:} \\
\{task\_description\}

\medskip

\textbf{DATA SUMMARIES:} \\
\{file\_summaries\_text\}

\medskip

Based on the provided task description and data summaries, please analyze the data and provide guidelines for simulation model construction.

\medskip

Your analysis must cover:
{\footnotesize
\begin{enumerate}\itemsep2pt
    \item \textbf{Overall simulation design}: State the primary objective of the simulation. Specify how the simulation initializes from inputs. Specify the execution of the simulation (should involve tuning parameters through data calibration). Specify what artifacts it outputs after execution (should be the calibrated parameters and the evaluation results on the validation dataset).
    \item \textbf{Scale \& Granularity}: Specify time step, spatial resolution (or explicitly "non-spatial"), and population size (agents), with brief rationale.
    \item \textbf{Agent Archetypes}: Define the agent unit and roles; list static attributes and dynamic states; explain how the input data construct static attributes and how dynamic states update from data-derived signals.
    \item \textbf{Interaction Topology}: Describe how agents interact; explain how to build interactions (layers/edges/protocols) from the input data.
    \item \textbf{Information propagation}: Clarify whether information diffusion exists, its topology/mechanism, and how to parameterize/drive it using inputs.
    \item \textbf{Exogenous Signals}: Identify any external signals/interventions, how they are derived from inputs, and how they affect agent decisions.
    \item \textbf{Action Decision Policy}: Describe actions taken by agents and the decision policy mapping observations/signals to actions; include role-specific inputs and policy forms.
    \item \textbf{Holdout}: Propose a training / validation data split plan if needed. For time series, prefer first 80\% of days as train and last 20\% as validation; state exact ranges or the rule to compute them.
    \item \textbf{Simulation Evaluation}: Define evaluation metrics and how to compare simulator output artifacts against validation ground truth.
\end{enumerate}
}

\medskip

Provide your response in the following JSON format (valid JSON only, no extra text):

{\scriptsize
\begin{lstlisting}[breaklines=true, breakatwhitespace=true, columns=fullflexible, frame=none]
{
  "overall_simulation_design": {
    "objective": "what is this simulation about",
    "initialization": {
      "description": "How the simulation starts (states, seeds, signals)",
      "source_type":
            "data_derived|designed|"
            "mixed|unknown"
      "data_reference": {"file": "filename", "fields": ["field_a"], "derivation": "how derived"}
    },
    "execution": "How to tune parameters through data calibration",
    "outputs": ["calibrated_parameters", "evaluation_results_on_validation"]
  },
  "scale_granularity": {
    "time_step": {"value": "seconds|minutes|hours|days", 
    "source_type":
  "data_derived|designed|"
  "mixed|unknown" 
    "data_reference": null},
    "spatial_resolution": {"value": "non-spatial|grid|POI|road network|other:<...>", "source_type": "data_derived|designed|"
        "mixed|unknown", "data_reference": null},
    "population_size": {"value": "integer or description tied to data", "source_type": "data_derived|designed|"
        "mixed|unknown", "data_reference": null},
    "rationale": "Why these scales are appropriate"
  },
  "agent_archetypes": {
    "unit": 
        "simulated_entity|user|"
        "household|device",
    "roles": [
      {
        "name": "role_name_derived_from_task",
        "static_attributes": [
          {
            "name": "attr_name",
            "type": 
                    "int|float|string|"
                    "enum|vector|other",
            "source_type": 
                    "data_derived|designed|"
                    "mixed|unknown",
            "data_reference": {"file": "filename", "fields": ["field_name"], "derivation": "mapping/aggregation rule"}
          }
        ],
        "dynamic_states": [
          {
            "name": "state_name",
            "type": 
                "int|float|string|"
                "enum|vector|other",
            "update_rule": "how it updates",
            "source_type": 
                "data_derived|designed|"
                "mixed|unknown",
            "data_reference": {"file": "filename", "fields": ["field_name"], "derivation": "mapping/aggregation rule"}
          }
        ],
        "construction_from_data": "Explain construction; if designed, explain assumption",
        "update_from_data": "Explain updates; if designed, explain assumption"
      }
    ]
  },
  "interaction_topology": {
    "topology": 
        "graph|hybrid|"
        "broadcast|platform-level",
    "layers": [
      {
        "name": "layer_name",
        "from_role": "role_name",
        "to_role": "role_name",
        "edge_rule": "how nodes/edges/events are formed",
        "source_type": 
            "data_derived|designed|"
            "mixed|unknown",
        "data_reference": {"file": "filename", "fields": ["src_id", "dst_id"], "derivation": "edge construction rule"}
      }
    ],
    "protocol": "describe message passing/order among these roles"
  },
  "information_propagation": {
    "exists": true,
    "topology": "which roles/channels are used for diffusion",
    "mechanism": "how diffusion works",
    "source_type": 
        "data_derived|designed|"
        "mixed|unknown",
    "data_reference": {"file": "filename", "fields": ["field_name"], "derivation": "drive intensity/schedule"}
  },
  "exogenous_signals": [
    {
      "name": "signal_name",
      "effect_on_agents": "how it enters decision function",
      "bounds": "[low, high] or rationale",
      "source_type": 
        "data_derived|designed|"
        "mixed|unknown",
      "data_reference": {"file": "filename", "fields": ["field_name"], "derivation": "how derived"}
    }
  ],
  "action_decision_policy": {
    "by_role": [
      {
        "role": "role_name_derived_from_task",
        "inputs": [
          {
            "name": "obs_or_signal_name",
            "source_type": 
                "data_derived|designed|"
                "mixed|unknown",
            "data_reference": {"file": "filename", "fields": ["field_name"], "derivation": "how computed"}
          }
        ],
        "policy_form": "policy description",
        "parameters": ["list of per-role parameter names (if any)"]
      }
    ]
  },
  "capability_realization": [
    {
      "role": "role_name_derived_from_task",
      "modes": ["heuristic_rules", "tool_calls", "llm_calls"],
      "llm_prompt_skeleton": "If llm_calls is used, provide a high-level prompt template with placeholders",
      "tool_dependencies": ["list of tool/data lookup functions (only cite files/fields when data-derived)"],
      "fallback_strategy": "what to do if LLM/tool is unavailable",
      "logging": "what intermediate results are written back to shared memory/log"
    }
  ],
  "holdout_plan": {
    "method": 
        "temporal_holdout|random_split|"
        "rolling_backtest",
    "time_ordering": {
      "source_type": 
        "data_derived|designed|"
        "mixed|unknown",
      "data_reference": {"file": "filename", "fields": ["time_field"], "derivation": "ordering rule"}
    },
    "train_range": "rule to compute train set",
    "validation_range": "rule to compute validation set",
    "notes": "split notes"
  },
  "simulation_evaluation": {
    "metrics": [
      {
        "name": "metric_name",
        "definition": "how to compute it",
        "ground_truth": {
          "source_type": 
            "data_derived|designed|"
            "mixed|unknown",
          "data_reference": {"file": "filename", "fields": ["target_field"], "derivation": "target extraction"}
        }
      }
    ],
    "comparison_method": "how to compute metrics on validation set and report"
  },
  "calibratable_parameters": [
    {
      "name": "parameter_name",
      "range_bounds": "[low, high] or rationale",
      "source": "constrained by data or modeling",
      "notes": "tie to interaction or decision speed"
    }
  ],
  "llm_and_tool_specs": {
    "llm_required": true,
    "llm_calls": [
      {
        "name": "llm_call_name_derived_from_role",
        "inputs": ["list of fields/placeholders to inject"],
        "output": "expected JSON/structured output"
      }
    ],
    "tool_wrappers": [
      {
        "name": "tool_function_name",
        "input_type": "string|object",
        "output_type": "structured_json"
      }
    ]
  }
}
\end{lstlisting}
}

Return only valid JSON that can be parsed. Do not include any other explanation or text outside the JSON.
\end{tcolorbox}

\begin{center}
\footnotesize
\textbf{Prompt:} \textit{This is the prompt of the \textbf{Data Analysis Agent} for generating blueprint $\mathcal{B}$.}
\label{prompt:data_analysis}
\end{center}

\subsection{Blueprint Example}
\label{ssec:blueprint_example}

In this section, we provide an illustrative example of a complete blueprint generated by the \textbf{Data Analysis Agent} for the \textbf{Mob. N$\to$A} scenario. This structured JSON object contains all the necessary information, from data analysis summaries to simulation design parameters, required by downstream agents to construct and execute the simulator.

\begin{tcolorbox}[
  enhanced,
  breakable,
  width=\linewidth,
  colback=gray!10,
  colframe=red!70!black,
  title={\textbf{Blueprint for Mob. N->A}},
  fonttitle=\bfseries,
  boxrule=0.8pt,
  arc=0pt,
  left=3pt,
  right=3pt,
  top=3pt,
  bottom=3pt,
  boxsep=2pt,
  before upper={\setlength{\parindent}{0pt}\ttfamily\footnotesize}
]

{\scriptsize
\begin{lstlisting}[breaklines=true, breakatwhitespace=true, columns=fullflexible, frame=none]
{
  "data_analysis_result": {
    "overall_simulation_design": {
      "objective": "Generate realistic daily mobility trajectories (ordered sequences of (POI, time) visits) for urban residents under an out-of-distribution Normal-to-Abnormal (N2A) setting: fit baseline behavior using only 2019\u20132020 records, tune/calibrate abnormal-shift parameters using a 2021 validation split, and report final performance on a held-out 2021 test split.",
      "initialization": {
        "description": ("Load and parse 1921Y.json into per-user, per-day sequences; split by year into normal (2019\u20132020) vs abnormal (2021). Build POI catalog with lat/lon from poi_category_192021_",
        "longitude_latitude.json and map fine categories to super-categories using catto.json. Initialize each Resident with baseline (normal-period) learned distributions (time-of-day, chain length, category/POI transitions, anchor POIs). Initialize global abnormal-shift parameters (to be calibrated) that perturb baseline distributions when simulating 2021."),
        "source_type": "mixed",
        "data_reference": {
          "file": "1921Y.json",
          "fields": [
            "top-level keys (user IDs)",
            "top-level values (daily activity strings containing dates, locations, times)"
          ],
          "derivation": ("Parse each string 'Activities at YYYY-MM-DD: ...' into date and ordered visits of (Category#POI_ID, HH:MM:SS); year determines normal (2019\u20132020) vs abnormal (2021). Join POI lat/lon by matching POI identifier strings to poi_category_192021_",
          "longitude_latitude.json entries; map Category -> super-category via catto.json when available.")
        }
      },
      "execution": "1) Fit baseline per-user and global mobility models on 2019\u20132020 only (e.g., distributions over start time, stop count, inter-visit time gaps, super-category transitions, POI choice kernels conditioned on category and distance). 2) Define abnormal-shift mechanisms with free parameters (e.g., category-mix reweighting, increased/decreased mobility radius, increased infrastructure/transit propensity, time-shift of activities). 3) Calibrate abnormal-shift parameters by simulating 2021 validation days and optimizing an objective over evaluation metrics (trajectory similarity, distributional match) against 2021 ground truth. 4) Freeze baseline + calibrated shift and evaluate on 2021 held-out test split. 5) Online OOD context window rule: For each 2021 target day, construct a 7-day context window using only days strictly earlier than the target day; this window may include earlier 2021 days and may backfill from late 2020; never use target-day or future days.",
      "outputs": [
        "calibrated_parameters",
        "evaluation_results_on_validation"
      ]
    },
    "scale_granularity": {
      "time_step": {
        "value": "minutes",
        "source_type": "mixed",
        "data_reference": {
          "file": "1921Y.json",
          "fields": [
            "visit time tokens HH:MM:SS within daily strings"
          ],
          "derivation": "Observed times are second-resolved; simulator can operate in minute granularity while outputting HH:MM:SS by rounding/stochastic seconds jitter (designed)."
        }
      },
      "spatial_resolution": {
        "value": "POI",
        "source_type": "data_derived",
        "data_reference": {
          "file": "1921Y.json",
          "fields": [
            "place tokens formatted Category#POI_ID"
          ],
          "derivation": "Each visit location is a POI identifier string (Category#id)."
        }
      },
      "population_size": {
        "value": "number of unique user IDs in 1921Y.json (one agent per key)",
        "source_type": "data_derived",
        "data_reference": {
          "file": "1921Y.json",
          "fields": [
            "top-level keys"
          ],
          "derivation": "Count keys in the JSON object; each key defines one resident agent."
        }
      },
      "rationale": "Trajectories are day-level sequences with intra-day timestamps; minute-level simulation supports feasibility constraints via travel-time approximations while matching the dataset\u2019s timestamp structure. POI-level resolution is required because ground truth uses Category#POI_ID and POI lat/lon is available for distance-based plausibility."
    },
    "agent_archetypes": {
      "unit": "resident/user",
      "roles": [
        {
          "name": "Resident",
          "static_attributes": [
            {
              "name": "resident_id",
              "type": "string",
              "source_type": "data_derived",
              "data_reference": {
                "file": "1921Y.json",
                "fields": [
                  "top-level key"
                ],
                "derivation": "Use JSON key as resident_id."
              }
            },
            {
              "name": "anchor_pois",
              "type": "vector",
              "source_type": "mixed",
              "data_reference": {
                "file": "1921Y.json",
                "fields": [
                  "daily visit sequences (locations)"
                ],
                "derivation": "From 2019\u20132020 visits, compute top-K frequent POIs; additionally mark any POI whose category is 'Home' as an anchor when present (data-derived), otherwise rely on frequency-only heuristic (designed fallback)."
              }
            },
            {
              "name": "baseline_start_time_dist",
              "type": "other",
              "source_type": "data_derived",
              "data_reference": {
                "file": "1921Y.json",
                "fields": [
                  "first timestamp per day"
                ],
                "derivation": "For each 2019\u20132020 day, extract first visit time; fit a distribution per user (e.g., histogram/mixture)."
              }
            },
            {
              "name": "baseline_stop_count_dist",
              "type": "other",
              "source_type": "data_derived",
              "data_reference": {
                "file": "1921Y.json",
                "fields": [
                  "number of visits per day"
                ],
                "derivation": "For each 2019\u20132020 day, count visits; fit a per-user distribution."
              }
            },
            {
              "name": 
                ("baseline_inter_event",
                "_time_dist"),
              "type": "other",
              "source_type": "data_derived",
              "data_reference": {
                "file": "1921Y.json",
                "fields": [
                  "ordered visit times within day"
                ],
                "derivation": "Compute successive time gaps within 2019\u20132020 days; fit distribution (conditioned on previous/next category optionally)."
              }
            },
            {
              "name":
                "baseline_category_"
                "transition_model",
              "type": "other",
              "source_type": "mixed",
              "data_reference": {
                "file": "1921Y.json",
                "fields": [
                  "ordered visit locations (Category#POI_ID)"
                ],
                "derivation": "Extract Category from each location token; optionally map to super-category via catto.json; fit Markov transition probabilities with smoothing at super-category level if category not found in catto.json."
              }
            },
            {
              "name": "baseline_poi_choice_model",
              "type": "other",
              "source_type": "mixed",
              "data_reference": {
                "fields": [
                  "POI records [lat, lon, 'Category#id']"
                ],
                "derivation": "Given chosen category, choose POI via a combination of (a) empirical POI popularity from 2019\u20132020 visits (data-derived from 1921Y.json) and (b) distance-decay from current POI using lat/lon (data-derived). Weighting between (a) and (b) is calibratable (mixed)."
              }
            }
          ],
          "dynamic_states": [
            {
              "name": "current_day",
              "type": "string",
              "update_rule": "Advance to next simulation date; for calibration/evaluation, simulate only dates in the selected 2021 validation/test split; for generation, dates are sampled from target set.",
              "source_type": "mixed",
              "data_reference": {
                "file": "1921Y.json",
                "fields": [
                  "date prefix 'Activities at YYYY-MM-DD'"
                ],
                "derivation": "Use available dates to define candidate simulation days (data-derived); selecting which days to simulate per split is designed by holdout plan."
              }
            },
            {
              "name": "current_poi",
              "type": "string",
              "update_rule": "Set to the most recently generated POI in the trajectory sequence.",
              "source_type": "designed",
              "data_reference": null
            },
            {
              "name": "current_time",
              "type": "string",
              "update_rule": "Set to generated time-of-day; increment by sampled inter-event gap subject to feasibility constraints.",
              "source_type": "mixed",
              "data_reference": {
                "file": "1921Y.json",
                "fields": [
                  "HH:MM:SS tokens"
                ],
                "derivation": "Time-of-day distributions fit from 2019\u20132020; feasibility adjustment (e.g., minimum travel time) is designed using POI distances."
              }
            },
            {
              "name": "daily_plan",
              "type": "vector",
              "update_rule": "At day start, sample stop_count, start_time, and a sequence of activity categories; then instantiate POIs and times sequentially.",
              "source_type": "mixed",
              "data_reference": {
                "file": "1921Y.json",
                "fields": [
                  "per-day sequences of categories and times"
                ],
                "derivation": "Plan templates and distributions learned from normal data; abnormal adjustments applied via exogenous shift parameters calibrated on 2021 validation."
              }
            },
            {
              "name": "visited_sequence",
              "type": "vector",
              "update_rule": "Append each generated (POI, time) as simulation progresses.",
              "source_type": "designed",
              "data_reference": null
            }
          ],
          "construction_from_data": ("Create one Resident agent per user ID in 1921Y.json. Parse 2019\u20132020 days to estimate baseline distributions and transition models. Build POI lookup table from poi_category_192021_",
          "longitude_latitude.json by indexing on the POI identifier string. Optionally map fine categories to super-categories using catto.json for smoothing and OOD robustness."),
          "update_from_data": "During calibration/evaluation, for a given 2021 day the agent generates a trajectory using baseline models plus abnormal-shift parameters. No online learning from 2021 ground truth during simulation; only outer-loop calibration uses 2021 validation targets to adjust global shift parameters."
        },
        {
          "name": "CalibrationController",
          "static_attributes": [
            {
              "name": "search_strategy",
              "type": "enum",
              "source_type": "designed",
              "data_reference": null
            },
            {
              "name": "objective_weights",
              "type": "vector",
              "source_type": "designed",
              "data_reference": null
            }
          ],
          "dynamic_states": [
            {
              "name": "current_parameter_vector",
              "type": "vector",
              "update_rule": "Update via black-box optimization (e.g., Bayesian optimization / evolutionary search / random search) to minimize validation loss computed against 2021 validation set.",
              "source_type": "designed",
              "data_reference": null
            },
            {
              "name": "best_parameter_vector",
              "type": "vector",
              "update_rule": "Replace when validation objective improves.",
              "source_type": "designed",
              "data_reference": null
            },
            {
              "name": "experiment_log",
              "type": "vector",
              "update_rule": "Append each trial\u2019s params, seed, metrics, and artifacts.",
              "source_type": "designed",
              "data_reference": null
            }
          ],
          "construction_from_data": "Controller reads parsed datasets and defines train/validation/test partitions per holdout plan. It does not exist in the raw data; it is introduced to meet the requirement of calibrating using 2021 validation while fitting baseline on 2019\u20132020.",
          "update_from_data": "Uses 2021 validation ground truth from 1921Y.json to compute objective; updates parameter vector based on optimization routine."
        }
      ]
    },
    "interaction_topology": {
      "topology": "platform-level",
      "layers": [
        {
          "name": "calibration_loop",
          "from_role": "CalibrationController",
          "to_role": "Resident",
          "edge_rule": "Controller broadcasts a candidate abnormal-shift parameter vector and simulation configuration (seed, day subset); Residents simulate trajectories for the requested days; outputs are returned for scoring.",
          "source_type": "designed",
          "data_reference": null
        },
        {
          "name": "shared_environment_lookup",
          "from_role": "Resident",
          "to_role": "Resident",
          "edge_rule": "No direct resident-to-resident influence; all residents share the same POI catalog, category taxonomy, and global shift parameters.",
          "source_type": "designed",
          "data_reference": null
        }
      ],
      "protocol": "Per calibration iteration: (1) Controller selects parameter vector \u03b8 and simulation seed(s). (2) For each resident/day in the calibration subset, residents generate trajectories using \u03b8. (3) Controller computes metrics vs 2021 ground truth and updates \u03b8. After tuning, run evaluation on held-out 2021 test split and report metrics."
    },
    "information_propagation": {
      "exists": false,
      "topology": "none (no explicit social diffusion in provided data)",
      "mechanism": "Residents do not exchange messages; any global changes are modeled via exogenous abnormal-shift parameters applied uniformly or stratified by resident clusters (optional, designed).",
      "source_type": "data_derived",
      "data_reference": {
        "file": "1921Y.json",
        "fields": [
          "user-indexed independent daily trajectories"
        ],
        "derivation": "Data contains per-user sequences without any explicit social links or interactions; therefore diffusion is not directly supported."
      }
    },
    "exogenous_signals": [
      {
        "name": "abnormal_period_shift",
        "effect_on_agents": "Applies when simulating dates in 2021: reweights category/super-category choice, modifies stop-count distribution, shifts start times, adjusts travel-radius/distance-decay, and adjusts propensity to include infrastructure categories (e.g., Toll Booth/Tunnel/Rest Area/Platform) as intermediate stops.",
        "bounds": "Calibration bounds set to reasonable multiplicative/shift ranges around baseline (e.g., category weight multipliers in [0.25, 4], time shifts in [-120,+120] minutes, distance-decay scale multipliers in [0.5, 2.5]).",
        "source_type": "mixed",
        "data_reference": {
          "file": "1921Y.json",
          "fields": [
            "2021 daily trajectories (locations and times)"
          ],
          "derivation": "Signal is designed as a parameterized mechanism; its parameters are calibrated to match distributional properties of 2021 trajectories."
        }
      },
      {
        "name": "day_type_proxy",
        "effect_on_agents": "Optional: adjust schedule templates for weekday/weekend using only date-derived heuristics; used to condition start time and stop-count distributions.",
        "bounds": "Binary {weekday, weekend} inferred from date; if timezone/calendar assumptions required, treat as heuristic with standard Gregorian calendar.",
        "source_type": "designed",
        "data_reference": null
      }
    ],
    "action_decision_policy": {
      "by_role": [
        {
          "role": "Resident",
          "inputs": [
            {
              "name":
                    "baseline_"
                    "distributions_models",
              "source_type": "data_derived",
              "data_reference": {
                "file": "1921Y.json",
                "fields": [
                  "2019\u20132020 parsed daily sequences"
                ],
                "derivation": "Fit per-user and global models from normal period only."
              }
            },
            {
              "name": "poi_catalog_with_latlon",
              "source_type": "data_derived",
              "data_reference": {
                "file": 
                ("poi_category_192021_",
                "longitude_latitude.json"),
                "fields": [
                  "category -> list of [lat, lon, poi_id]"
                ],
                "derivation": "Index POIs by poi_id and by category; compute distances as needed."
              }
            },
            {
              "name":
                "category_to_"
                "supercategory_map",
              "source_type": "data_derived",
              "data_reference": {
                "file": "catto.json",
                "fields": [
                  "category -> super-category"
                ],
                "derivation": "Lookup for each fine category; if missing, treat super-category as 'Unknown' (designed fallback)."
              }
            },
            {
              "name":
                "abnormal_period_"
                "shift_parameters",
              "source_type": "designed",
              "data_reference": null
            },
            {
              "name": "current_state (current_poi, current_time, visited_sequence)",
              "source_type": "designed",
              "data_reference": null
            }
          ],
          "policy_form": "Hierarchical generative policy per day: (1) Sample start time and stop count from baseline distributions perturbed by abnormal shift (if 2021). (2) Sample a sequence of activity categories using a Markov model (category or super-category) with optional time-of-day conditioning and abnormal reweighting. (3) For each category, sample a concrete POI using a mixture of (a) personal historical preference (2019\u20132020 POI visit frequency) and (b) spatial distance-decay from current POI using lat/lon; optionally enforce feasibility by ensuring time gaps exceed a minimum travel-time proxy derived from distance. (4) Output ordered (POI, time) visits.",
          "parameters": [
            "theta_cat_weight_multiplier_"
            "by_supercategory",
            "theta_stop_count_multiplier",
            "theta_start_time_shift_minutes",
            "theta_distance_decay_scale",
            "theta_preference_vs_"
            "distance_mixture",
            "theta_infrastructure_stop_bonus"
          ]
        },
        {
          "role": "CalibrationController",
          "inputs": [
            {
              "name": "simulated_trajectories",
              "source_type": "designed",
              "data_reference": null
            },
            {
              "name": 
                "validation_ground_"
                "truth_2021",
              "source_type": "data_derived",
              "data_reference": {
                "file": "1921Y.json",
                "fields": [
                  "2021 daily activity strings"
                ],
                "derivation": "Parse 2021 dates/visits for validation subset."
              }
            }
          ],
          "policy_form": "Outer-loop calibration policy: propose \u03b8, run simulation on 2021 validation subset, compute objective from metrics, update \u03b8 using selected black-box optimization strategy.",
          "parameters": [
            "optimizer_hyperparams",
            "metric_weights"
          ]
        }
      ]
    },
    "holdout_plan": {
      "method": "temporal_holdout",
      "time_ordering": {
        "source_type": "data_derived",
        "data_reference": {
          "file": "1921Y.json",
          "fields": [
            "date token in each daily record string: 'Activities at YYYY-MM-DD'"
          ],
          "derivation": "Parse YYYY-MM-DD and sort chronologically; year used for regime split."
        }
      },
      "train_range": "Fit baseline only on dates with year in {2019, 2020}. Within 2021, do not use for baseline fitting.",
      "validation_range": "Within 2021 dates, use first 80% of dates in chronological order as validation for abnormal-shift calibration (per user or globally; implement globally by date ordering). (FIXED SPLIT RULE: per-user-by-date only) For each resident independently, sort that resident's 2021 day-records by date ascending; assign the first 80% of that resident's 2021 records to V_calib and the last 20% to V_test (deterministic, no global mixing). If a resident has fewer than 5 total 2021 records, exclude that resident entirely from calibration/test with explicit logging (default), or assign all their 2021 records to V_test only (allowed fallback) \u2014 but never include them in V_calib.",
      "notes": "Final test is the remaining last 20% of 2021 dates chronologically. If some users have sparse 2021 records, apply the 80/20 rule per-user to preserve personalization (designed choice); record which approach is used. Split strategy is fixed to per-user-by-date as specified above; do not use global-by-date splitting."
    },
    "simulation_evaluation": {
      "metrics": [
        {
          "name": "stop_count_mae",
          "definition": "Mean absolute error between simulated and ground-truth number of visits per (user, day).",
          "ground_truth": {
            "source_type": "data_derived",
            "data_reference": {
              "file": "1921Y.json",
              "fields": [
                "parsed 2021 daily visit sequences"
              ],
              "derivation": "Count visits in each 2021 day string."
            }
          }
        },
        {
          "name": "category_mix_jsd",
          "definition": "Jensen-Shannon divergence between simulated vs ground-truth distributions of fine categories (or super-categories) aggregated over the validation set.",
          "ground_truth": {
            "source_type": "mixed",
            "data_reference": {
              "file": "1921Y.json",
              "fields": [
                "Category#POI_ID tokens in 2021 days"
              ],
              "derivation": "Extract Category from tokens; optionally map to super-category via catto.json when computing super-category mix."
            }
          }
        },
        {
          "name": "time_of_day_emd",
          "definition": "Earth Mover's Distance between simulated vs ground-truth distributions of visit times (e.g., histogram over minutes-of-day) aggregated over validation set.",
          "ground_truth": {
            "source_type": "data_derived",
            "data_reference": {
              "file": "1921Y.json",
              "fields": [
                "HH:MM:SS tokens in 2021 days"
              ],
              "derivation": "Convert to minutes-of-day; build histogram."
            }
          }
        },
        {
          "name": "poi_topk_recall",
          "definition": "For each (user, day): compute recall@K of ground-truth POIs appearing in the simulated day (set-based), average across days. K is fixed (e.g., 5 or length of simulated sequence, designed).",
          "ground_truth": {
            "source_type": "data_derived",
            "data_reference": {
              "file": "1921Y.json",
              "fields": [
                "POI identifier tokens Category#POI_ID in 2021 days"
              ],
              "derivation": "Extract POI identifier strings from parsed visits."
            }
          }
        },
        {
          "name": "radius_of_gyration_error",
          "definition": "Absolute/relative error between simulated and ground-truth daily radius of gyration computed from POI coordinates (requires POI lat/lon).",
          "ground_truth": {
            "source_type": "mixed",
            "data_reference": {
              "file": 
                ("poi_category_192021_",
              longitude_latitude.json"),
              "fields": [
                "POI records [lat, lon, poi_id]"
              ],
              "derivation": "Join 2021 visited POIs to coordinates; compute radius of gyration per day; handle missing POIs by excluding those visits (designed missingness handling)."
            }
          }
        }
      ],
      "comparison_method": "Compute metrics on the 2021 validation split for each calibration trial; select theta minimizing a weighted sum of metrics (objective). After calibration, compute the same metrics on the 2021 test split and report. Hard constraint: best_params selection and early stopping must use ONLY V_calib; V_test must be evaluated exactly once after selection and must not affect any choice (no tuning, no re-selection, no peeking)."
    },
    "calibratable_parameters": [
      {
        "name":
            "theta_cat_weight_multiplier_"
            "by_supercategory",
        "range_bounds": "Multipliers per super-category in [0.25, 4.0]",
        "source": "constrained by modeling",
        "notes": "Reweights baseline category/super-category selection in 2021 to match abnormal mix."
      },
      {
        "name": "theta_stop_count_multiplier",
        "range_bounds": "[0.5, 1.8]",
        "source": "constrained by modeling",
        "notes": "Scales expected number of daily stops in 2021 vs baseline."
      },
      {
        "name": "theta_start_time_shift_minutes",
        "range_bounds": "[-120, 120]",
        "source": "constrained by modeling",
        "notes": "Shifts start-time distribution in 2021."
      },
      {
        "name": "theta_distance_decay_scale",
        "range_bounds": "[0.5, 2.5]",
        "source": "constrained by modeling",
        "notes": "Controls travel radius by scaling distance sensitivity in POI choice."
      },
      {
        "name": 
            "theta_preference_vs_"
            "distance_mixture",
        "range_bounds": "[0.0, 1.0]",
        "source": "constrained by modeling",
        "notes": "Mixture weight: 1.0=personal preference only; 0.0=distance-only within category."
      },
      {
        "name": "theta_infrastructure_stop_bonus",
        "range_bounds": "[0.0, 3.0]",
        "source": "constrained by modeling",
        "notes": "Boosts probability of selecting travel/infrastructure-like categories as intermediate stops in 2021."
      }
    ]
  }
}
\end{lstlisting}
}
\end{tcolorbox}

\begin{center}
\footnotesize
\textit{This is the blueprint example of the \textbf{Mob. N$\to$A} task.}
\label{example:blueprint}
\end{center}

\section{Instructions Given to Participants}
\label{sec:participant_instructions}

For the HITL Blueprint verification and focus-group evaluation, participants were provided with a document titled \textbf{Participant Information Statement and Consent Form}. The participant-facing content is reproduced below with minor adjustments for reporting.

\subsection{Participant Information Statement and Consent Form}
\label{ssec:piscf}

\paragraph{Participant Group:} Domain Experts and Social Scientists

\paragraph{Study Title:} \textit{\socia: Automated Simulator Construction via Dual-Anchored Bi-Level Optimization}

\paragraph{Chief Investigator:} Flora Salim

\paragraph{What is the research study about?}
You are invited to participate in a research study exploring how large language model (LLM) agents can be used to automatically generate high-fidelity social simulations in urban planning, healthcare, and digital platforms. These simulations are built using the \socia framework and serve as interactive environments to test policy decisions, service design, and sociotechnical interventions. We aim to collect your feedback as an expert or user to evaluate these simulations---either through a survey, an interview, or a focus group discussion---and guide future refinement of the \socia system.

\paragraph{Research Funder:} ARC Centre of Excellence for Automated Decision Making and Society (CE200100005).

\paragraph{Inclusion/Exclusion Criteria}
Before you decide to participate in this research study, we need to ensure that it is appropriate for you to take part.

\paragraph{Inclusion criteria:}
\begin{itemize}
    \item Be aged 18 years or older;
    \item Be fluent in English, determined by either:
    \begin{itemize}
        \item completion of a tertiary degree in English, or
        \item regular professional use of English in workplace communication;
    \end{itemize}
    \item Hold at least a bachelor’s degree.
\end{itemize}

\paragraph{Exclusion criteria:}
\begin{itemize}
    \item Do not meet all of the inclusion criteria listed above;
    \item Are directly involved in the development or implementation of the \socia system or related simulation tools, to avoid biased feedback.
\end{itemize}

\paragraph{Do I have to take part in this research study?}
Participation in this research study is voluntary. If you do not want to take part, you do not have to. If you decide to take part and later change your mind, you are free to withdraw from the study at any stage.

If you decide you want to take part in the research study, you will be asked to:
\begin{itemize}
    \item read the information carefully (and ask questions if necessary);
    \item sign the consent form if you decide to participate in the study.
\end{itemize}

\paragraph{What does participation in this research require, and are there any risks involved?}
You may skip any question or withdraw at any time. Participation is voluntary. Both the online and offline focus groups will be recorded---with your consent---for transcription.
While participation in this study is considered low risk, there are still potential risks that participants should be aware of:

\paragraph{Psychological discomfort.}
Some participants may feel mild stress, discomfort, or frustration when using the \socia system, particularly if they are unfamiliar with simulations, coding, or interpreting model outputs. Focus group discussions may also involve critical reflection on policy or societal issues, which could cause discomfort in expressing personal opinions.

\begin{itemize}
    \item \textbf{Likelihood:} Possible but unlikely.
    \item \textbf{Severity:} Mild, typically limited to momentary stress, frustration, or discomfort during simulation use or group discussion.
    \item \textbf{Management:} Participants will be reminded that they may skip any question or stop using the system at any time without consequence. Focus groups will be moderated by facilitators who will monitor participants’ wellbeing and provide breaks if needed.
\end{itemize}

\paragraph{Privacy and confidentiality.}
Participation involves the collection of demographic information, professional background, and discussion data during focus groups. Recordings (audio/video) may contain identifiable information.
\begin{itemize}
    \item \textbf{Likelihood:} Very low, as strict data protection protocols will be applied.
    \item \textbf{Severity:} Moderate if breaches occur, as personal opinions or professional background could be linked to individuals.
    \item \textbf{Management:} All data will be securely stored on a UNSW password-protected OneDrive and only accessible to the approved research team. Any publication of results will use aggregated or anonymised data to prevent identification. After transcription and re-identification, all video and audio recordings will be permanently deleted. Only the re-identified transcripts will be retained for analysis and reporting.
\end{itemize}

\paragraph{Technical risks.}
Participants may experience technical difficulties (e.g., unstable internet connection, difficulties using the \socia interface).
\begin{itemize}
    \item \textbf{Likelihood:} Moderate, given reliance on online platforms. 
    \item \textbf{Severity:} Mild, limited to inconvenience or disruption of participation. 
    \item \textbf{Management:} Technical support will be available before and during the sessions. If technical issues occur, participants can ask facilitators for assistance.
\end{itemize}

\paragraph{Time burden.}
The study requires approximately 60--90 minutes for the focus group and a few minutes for pre- and post-surveys.
\begin{itemize}
    \item \textbf{Likelihood:} Certain. 
    \item \textbf{Severity:} Mild, representing inconvenience only. 
    \item \textbf{Management:} To acknowledge this time commitment, participants will be reimbursed with a Coles \& Myer gift card at the standard Australian hourly rate, estimated at AUD 20 per hour.
\end{itemize}

\paragraph{Screening}
We ask that you complete a short pre-focus group survey about your academic or professional background and your experience with simulations; this will determine if you are eligible to participate. Completing the screening measures will take approximately 5 minutes.

\paragraph{Attendance modalities}
\begin{itemize}
    \item \textbf{Online Focus Group.}
These sessions take place via Teams or Zoom and last approximately 60--90 minutes. You will be informed of the date and provided with an invitation once we receive this consent form and your screening responses. During the focus group, you will first be introduced to how \socia operates as a platform. You will then be asked to use the platform by typing your desired simulation scenarios in natural language into our program. This will create code to run statistical models and provide visual outputs in the form of graphs, diagrams, and tables. After interacting with \socia, we will hold small group discussions, broadly grouped by participants with and without coding backgrounds. These discussions will be guided by our questions, with a few additional coding-specific questions for participants with coding experience. You will be asked to discuss your interpretation of the outcomes and your suggestions for improvement. This will be followed by a short post-focus group survey. The session will be recorded and transcribed to analyse and improve the \socia platform.
    \item \textbf{In-Person Focus Group.}
These sessions also last approximately 60--90 minutes and take place at UNSW at the ADM+S Symposium. Participants will be informed of the date and provided with a Teams invite for event communication once we receive the consent form and screening questionnaire. The session format is the same: an introduction to \socia, participant interaction with the platform by entering simulation scenarios in natural language, visual outputs generated by the system, and small-group discussions about interpretation and suggestions for improvement, followed by a short post-focus group survey.
\end{itemize}

\paragraph{Additional Costs and Reimbursement}
There are no costs associated with participating in this research project. While you will not receive payment, you will be provided with a prepaid Coles and Myer store gift card as reimbursement for your time and contribution. The reimbursement will be calculated at the standard Australian hourly rate, estimated at AUD 20 per hour, in line with typical research participation practices and the Fair Work Ombudsman’s guidance for casual research-related tasks. The reimbursement amount is the same for all participants, regardless of technical or non-technical background or scenario type.

\paragraph{What will happen to information about me?}
By signing the consent form, you consent to the research team collecting and using information about you for this research study.

The research team will store the data collected from you for:
\begin{itemize}
    \item a minimum of 3 years after the publication of the research results.
\end{itemize}

Information about you will be stored in a re-identifiable format where identifiers such as your name will be replaced by unique codes. After transcription and re-identification, all video and audio recordings will be permanently deleted. Only the re-identified transcripts will be retained for analysis and reporting.

You will be asked to provide consent for future use of the information collected from you in research that remains specific to the aims of this study. Your information will only be shared in a format that will not identify you.

Electronic data and recordings will be stored on a UNSW password-protected OneDrive accessible only to the approved research investigators. Recordings will only be made available after a confidentiality agreement has been signed.

The information you provide is personal information for the purposes of the Privacy and Personal Information Protection Act 1998 (NSW). You have the right of access to personal information held about you by the University, the right to request correction and amendment, and the right to make a complaint about a breach of the Information Protection Principles.

\paragraph{How and when will I find out what the results of the research study are?}
The research team intends to publish and/or report the results of the research. All information will be published in a way that will not identify you.

If you would like to receive a copy of the results, you can let the research team know by inserting your email or mailing address in the consent form. These details will only be used to send you the research results.

\paragraph{What if I want to withdraw from the research study?}
If you consent to participate, you may withdraw at any time. You can contact the research team and tell them you no longer want to participate. Your decision not to participate or to withdraw from the study will not affect your relationship with UNSW Sydney or any of the organisations involved in this research.

If you decide to leave the research study, the researchers will not collect additional information from you. You can request that any identifiable information about you be withdrawn from the research project.

\paragraph{What if I have a complaint or any concerns about the research study?}
If you have a complaint regarding any aspect of the study or the way it is being conducted, you may contact the UNSW Human Research Ethics Coordinator.

\paragraph{What should I do if I have further questions about my involvement in the research study?}
\textbf{Research Team Contact}
If you require further information regarding this study or have any problems related to your involvement, you may contact our and Chief Investigator.

\paragraph{Consent Form --- Participant providing own consent}
Participants were asked to confirm the following:
\begin{itemize}
    \item I understand I am being asked to provide consent to participate in this research study;
    \item I have read the Participant Information Sheet, or someone has read it to me in a language that I understand;
    \item I understand the purposes, study tasks and risks of the research described in the study;
    \item I understand that the research team will audio/video record the focus groups, and I agree to be recorded for this purpose;
    \item I provide my consent for the information collected about me to be used for the purpose of this research study only;
    \item I have had an opportunity to ask questions and am satisfied with the answers I have received;
    \item I freely agree to participate in this research study and understand that I am free to withdraw at any time;
    \item I may request a copy of the study results via email or post.
\end{itemize}

\paragraph{Optional consent items}
Participants were also asked whether they consented to:
\begin{itemize}
    \item making the collected information available to other researchers as described in the document;
    \item being identified in publications relating to this research;
    \item having their name and contact details retained in a register so they could be contacted about other research projects in the future.
\end{itemize}

\paragraph{Withdrawal of participation}
The document also provided a withdrawal form allowing participants to:
\begin{itemize}
    \item withdraw consent to participate in the study;
    \item request withdrawal of identifiable information previously provided;
    \item withdraw from further components while permitting retention/use of already collected information where applicable;
    \item acknowledge that information already published or not linked to identity cannot be withdrawn.
\end{itemize}

\end{document}